\documentclass{article}

\usepackage[preprint]{neurips_2026}

\usepackage[utf8]{inputenc} % allow utf-8 input
\usepackage[T1]{fontenc}    % use 8-bit T1 fonts
\usepackage{hyperref}       % hyperlinks
\usepackage{url}            % simple URL typesetting
\usepackage{booktabs}       % professional-quality tables
\usepackage{amsfonts}       % blackboard math symbols
\usepackage{nicefrac}       % compact symbols for 1/2, etc.
\usepackage{microtype}      % microtypography
\usepackage{xcolor}         % colors

\usepackage{amsmath}
\usepackage{amssymb}
\usepackage{mathtools}
\usepackage{amsthm}
\usepackage{graphicx}
\usepackage{refcount}
\usepackage{makecell}

\usepackage{bm}
\usepackage{subcaption}

\renewcommand{\vec}[1]{\bm{#1}}

\newcommand\Tau{\mathcal{T}}

\theoremstyle{plain}
\newtheorem{theorem}{Theorem}[section]

\theoremstyle{definition}
\newtheorem{definition}[theorem]{Definition}
\newtheorem{assumption}[theorem]{Assumption}
\theoremstyle{remark}
\newtheorem{remark}[theorem]{Remark}

\title{Observable Neural ODEs for Identifiable Causal Forecasting in Continuous Time}

\author{%
	Jennifer Wendland\\
	\texttt{wendland@uni-koblenz.de}
	\And
	Nicolas Freitag\\
	\texttt{nfreitag@uni-koblenz.de}
	\And
	Maik Kschischo\\
	\texttt{kschischo@uni-koblenz.de}\\
	Department of Computer Science\\
	University of Koblenz\\
	53424 Koblenz, Germany \\
	\\
}

\begin{document}

\maketitle

\begin{abstract}

Causal inference in continuous-time sequential decision problems is challenged by hidden confounding and partially observed states. We show that, under explicit structural assumptions, observability of the latent state enables identification of dynamic treatment effects through a continuous-time conditional front-door adjustment, even in the presence of hidden confounding.

We derive a general adjustment formula and show that it reduces to a tractable state-space formula when unobserved contemporaneous disturbances are temporally uncorrelated. This formula expresses potential-outcome distributions under alternative treatment trajectories through the measurement model, latent dynamics, and the filtering distribution over latent states.

We propose Observable Neural ODEs (ObsNODEs), Neural ODE models in observable normal form that implement this tractable adjustment for causal forecasting. ObsNODEs learn continuous-time dynamics with states reconstructible from observations, enabling outcome prediction under alternative treatment paths.

Experiments on synthetic, semi-synthetic, and real-world clinical data demonstrate strong performance over recent sequence models, including external validation.
\end{abstract}

\section{Introduction}
Sequential decision-making requires predicting how actions affect future outcomes under evolving observations. In medicine, this arises in dynamic treatment regimes (DTRs), where treatments are repeatedly adapted to a patient's longitudinal history \citep{murphy_optimal_2003,robins_optimal_2004}.

A central challenge is the identification of causal treatment effects in continuous time. In observational data, time-dependent confounding and treatment--confounder feedback arise when covariates affect both treatment assignment and outcomes while themselves being affected by earlier treatments \citep{hernan_causal_2023}. Identification becomes even more difficult when relevant components of the evolving disease state are latent and hidden variables affect both treatment and outcomes.

We address this problem from a state-space perspective. Under explicit structural assumptions, we show that observability makes the latent state reconstructible from treatment-outcome histories and thereby enables identification through a continuous-time conditional front-door adjustment, even in the presence of hidden confounding. We derive a general adjustment formula and show that, for temporally uncorrelated contemporaneous disturbances, it reduces to a tractable state-space formula. Based on this result, we introduce \emph{ObsNODEs}, Neural ODE models in observable normal form for causal forecasting under discrete and continuous treatment trajectories.

\textbf{Contributions.}
(i) We link control-theoretic observability to causal identifiability in latent continuous-time state-space models and derive a general conditional front-door adjustment for dynamic treatment effects under hidden confounding.
(ii) We derive a tractable state-space adjustment under temporally uncorrelated contemporaneous disturbances and propose \emph{ObsNODEs}, which enforce observability by construction for causal forecasting under dynamic treatment regimes.
(iii) We demonstrate strong causal forecasting performance on synthetic, semi-synthetic, and real-world clinical data compared with recent sequence models, including external validation.

\section{Problem statement}
We assume hat we observe a continuous-time multivariate outcome process $\vec{Y}_t$ and a time-varying treatment process $\vec{A}_t$ for times  $t\in [0,t_c)$. Given the observed history $\{\vec{a}_{[0,t_c)},\vec{y}_{[0,t_c]}\}$ with $\vec{a}_{[0,t_c)}=\{\vec{A}_t=\vec{a}_t|0\le t<t_c\}$ and $\vec{y}_{[0,t_c]}=\{\vec{Y}_t=\vec{y}_t|0\le t\le t_c\}$, the goal is to predict the probability density of the conditional potential outcome $\vec{Y}_t(\vec{a}_{[t_c,t_c+s)})$ on $[t_c,t_c+s)$ under a hypothetical treatment path  $\vec{a}_{[t_c,t_c+s)}=\{\vec{A}_t=\vec{a}_t|t_c\le t<t_c+s\}$. More precisely, we seek to identify the conditional potential-outcome density (CPOD) $P_{\vec{Y}_{t_c+s}(\vec{a}_{[t_c,t_c+s)})|\vec{Y}_{t_c},\vec{O}_{[0,t_c)}} (\vec{y}_{t_c+s}|\vec{y}_{t_c},\vec{o}_{[0,t_c)})$, given the last observation  $\vec{y}_{t_c}$ and the history 
of observations and treatments $\vec{o}_{[0,t_c)}=\{\vec{a}_{[0,t_c)},\vec{y}_{[0,t_c)}\}$.

In clinical applications, $\vec{Y}_t$ may contain irregularly sampled laboratory values, vital signs, or imaging embeddings, while $\vec{A}_t$ represents binary, categorical, or continuous treatments. Outside randomized trials, treatment assignment depends on the evolving patient history, which induces time-dependent confounding and treatment-confounder feedback.

We treat all measured variables as components of $\vec{Y}_t$ and define outcomes, when needed, as functions of $\vec{Y}_t$. This entails no loss of generality for the identification results below. We use outcome and observed process interchangeably for $\vec{Y}_t$.

\section{Preliminaries}

\subsection{Nonlinear state-space model}

A continuous-time deterministic state-space model is given by
%\begin{subequations}
%	\label{eq:dssm}
%	\begin{align}
%		\dot{\vec{z}}_t &= \vec{f}(\vec{z}_t,\vec{a}_t),\\
%		\vec{y}_t &= \vec{h}(\vec{z}_t),\\
%		\vec{z}_0 &= \vec{\zeta},19 /
%	\end{align}
%\end{subequations}
\begin{equation}
\label{eq:dssm}
\begin{aligned}
    \dot{\vec{z}}_t &= \vec{f}(\vec{z}_t,\vec{a}_t)
    && \text{(1a)},
    &
    \vec{y}_t &= \vec{h}(\vec{z}_t)
    && \text{(1b)},
    &
    \vec{z}_0 &= \vec{\zeta}
    && \text{(1c)}
\end{aligned}
\end{equation}
where $\vec{z}_t\in\mathbb{R}^{d_z}$ is a latent state, $\vec{a}_t\in\mathbb{R}^{d_a}$ a treatment or control, and $\vec{y}_t\in\mathbb{R}^{d_y}$ the observed process. The measurement function $\vec{h}$ is not necessarily invertible, in particular when $d_z>d_y$.  We assume $\vec{f}$ is Lipschitz in $\vec{z}$, so the system admits a unique solution for any admissible control path and initial condition.

For a given control path, observability asks whether the latent state or, equivalently, its initial value, is uniquely determined by the observed output over a finite horizon \citep{bernard_observer_2019}. For a given control $\vec{a}$ and initial condition $\vec{z}_0=\vec{\zeta}$, let $\vec{z}_t(\vec{a},\vec{\zeta})$ denote the resulting state trajectory, and define the corresponding output trajectory by $\vec{y}_t(\vec{a},\vec{\zeta}) := \vec{h}\!\left(\vec{z}_t(\vec{a},\vec{\zeta})\right)$.

\begin{definition}\label{def:observability}
	The system \eqref{eq:dssm} is \textbf{uniformly observable} on an open set $U\subset\mathbb{R}^{d_z}$ if, for any control path $\vec{a}:[0,T]\to\mathbb{R}^{d_a}$ and any $\vec{\zeta},\vec{\eta}\in U$,
	\[
	\vec{y}_t(\vec{a},\vec{\zeta})=\vec{y}_t(\vec{a},\vec{\eta})\;\;\forall t\in[0,T]
	\quad\Longrightarrow\quad
	\vec{\zeta}=\vec{\eta},
	\]
	where $[0,T]$ is the maximal interval on which both solutions exist.
\end{definition}

\subsection{Potential outcomes and graphical causal models}

We use the potential-outcomes framework and graphical causal models to formalize interventions \citep{robins_new_1986,pearl_causality_2000,peters_elements_2017,hernan_causal_2023}. We assume consistency, the stable unit treatment value assumption (SUTVA), and positivity. SUTVA excludes interference between units and multiple versions of treatment \citep{rubin_randomization_1980,hernan_causal_2023}, while positivity requires that the considered interventions lie within the support of the observational treatment process conditional on the relevant history. In longitudinal observational settings, identifiability of causal treatment effects is complicated by time-dependent confounding and treatment-confounder feedback.

\subsubsection{The conditional front-door criterion}
To connect causal identification with latent state-space models, we use the conditional front-door criterion \citep{xu_causal_2024}, an extension of the front-door criterion \citep{pearl_causality_2000} for estimating the effect of an intervention in the presence of an unobserved common cause of both treatment and outcome:
\begin{theorem}\label{thm:CFD} \citep{xu_causal_2024}
	A set of variables $M$ fulfills the \textbf{conditional front-door criterion (CFD)} relative to the ordered pair of variables $(A,Y)$  in a directed acyclic graph (DAG) $\mathcal{G}$, if (i)~$M$ intercepts all directed paths from $A$ to $Y$, (ii)~there exists a set of conditioning variables $W$ that blocks all backdoor paths from $A$ to $M$ and (iii) all backdoor paths between $M$ and $Y$ are blocked by $\{A\} \cup W$. If $M$ fulfills the CFD relative to $(A,Y)$ with conditioning variables $W$ the causal effect estimate is given by
	\begin{equation}\label{eq:cfd_adjust}
		P(Y(a)) = \sum_{m,w,a'} P(y|a',m,w)P(a'|w)P(m|a,w)P(w),
	\end{equation}
	where the sum over $a'$ runs over alternative treatments.
\end{theorem}

\section{Observable Neural ODEs for dynamic treatment regimes}
\begin{figure}
	\centering
	\includegraphics[width=0.9\linewidth]{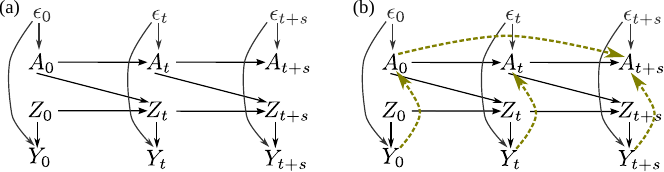}
	\caption{(a) Temporal DAG for treatment $A$, observation $Y$, and latent state $Z$ at three consecutive time points. The temporally independent latent disturbances $\epsilon$ jointly affect contemporaneous treatment assignment and observations. (b) Under a dynamic treatment regime, treatment additionally depends on the observed treatment-outcome history, as indicated by the green dashed arrows.}
	\label{fig:fig1dags}
\end{figure}
We model the observed process by a latent Neural ODE  \citep{chen_neural_2018,rubanova_latent_2019,kidger_neural_2020} with parameters $\theta$
\begin{align}\label{eq:node_ssm}
	\vec{Z}_0 \sim p_\theta(\vec{Z}_0),\qquad
	\dot{\vec{Z}}_t = \vec{f}_{\theta}(\vec{Z}_t,\vec{A}_t),\qquad
	\vec{Y}_t \sim q_\theta(\cdot|\vec{Z}_t),
\end{align}
where $p_\theta$ parameterizes the initial state distribution and $\vec{f}_\theta$ the vector field. The state is mapped via the dsitribution $q(\cdot|\vec{Z}_t,\vec{\epsilon}_t)$ to the observations and the observation model $q_\theta$ describes the marginal observation distribution. We interpret the treatment process as an intervention on the latent dynamics, as illustrated in Fig.~\ref{fig:fig1dags}(a). Typically, the treatment is adjusted according to the observations and we have a dynamic treatment regime~\citep{murphy_optimal_2003,robins_optimal_2004,hernan_causal_2023}: $\vec{A}_t = \vec{\pi}(\vec{y}_t,\vec{y}_{[0,t)},\vec{a}_{[0,t)})$ (see Fig.~\ref{fig:fig1dags}(b)). The temporally independent latent disturbance $\vec{\epsilon}_t$ may additionally affect both contemporaneous treatment assignment and observation, as illustrated in Fig.~\ref{fig:fig1dags}.

\subsection{Causal identifiability for the state-space model}
To establish an adjustment formula for the CPOD, we make the following assumptions 
for the data generating process, see Fig.~\ref{fig:fig1dags}.
\begin{assumption}\label{ass:latent}
    The latent state $\vec{Z}_t$ is a controlled continuous-time Markov process with  transition probability density $p(\vec{z}_{t+s}|\vec{a}_{[t,t+s)},\vec{z}_t)$.
\end{assumption}
\begin{assumption}\label{ass:observation}
    Given the latent state, the observation is independent of previous observations and treatments: 
    $p(\Vec{y}_t|\Vec{z}_t, \Vec{y}_{[0,t)},\Vec{a}_{[0,t)}) =  p(\Vec{y}_t|\Vec{z}_t)$.
\end{assumption}

\begin{theorem}\label{thm:obsid} Under consistency, SUTVA, overlap (positivity), Assumptions \ref{ass:latent} and \ref{ass:observation}, and observability of the latent state, the conditional potential-outcome distribution under $\vec a_{[t,t+s)}$ is identified by the adjustment formula
\begin{align}\label{eq:outcome_ident}
	&P_{\vec{Y}_{t+s}(\vec{a}_{[t,t+s)})|\vec{Y}_t,\vec{O}_{[0,t)}}
	(\vec{y}_{t+s}|\vec{y}_t,\vec{o}_{[0,t)}) \nonumber\\
	&\qquad=
	\int_{\vec{z}_{t+s}}\int_{\vec{z}_t}
	p(\vec{z}_t|\vec{y}_t,\vec{o}_{[0,t)})
	\,p(\vec{z}_{t+s}|\vec{a}_{[t,t+s)},\vec{z}_t)\,
	p(\vec{y}_{t+s}|\vec{z}_{t+s}),
\end{align}
This identification remains valid when temporally independent latent disturbances $\vec{\epsilon}_t$ induce unobserved contemporaneous dependence between treatment assignment and observations. The result also applies to dynamic treatment regimes under sequential positivity, with the transition density induced by the regime (see Fig.~\ref{fig:fig1dags}(b)).
\end{theorem}

\begin{remark}
Assumption~\ref{ass:observation} holds when $\vec{\epsilon}_t$ is an exogenous, temporally independent disturbance. With serially correlated hidden disturbances, the causal effect remains identifiable through the general conditional front-door adjustment, but the formula no longer reduces to \eqref{eq:outcome_ident}; see \eqref{eq:general_continuous_adjust} in Appendix~\ref{secapp:ident_derivation}.
\end{remark}
\begin{proof}\label{proof:adj_formula}
We derive a discrete-time version of the adjustment formula \eqref{eq:outcome_ident} in Theorem~\ref{thm:obsid} and relegate details for the continuous-time version to   
Appendix~\ref{secapp:ident_derivation}. Consider three consecutive time points $0,t$ and $t+s$.  A time-discretized approximation of the ODE in \eqref{eq:node_ssm} is given by
$\vec{Z}_{t+s} \approx \vec{Z}_t + \vec{f}(\vec{Z}_t,\vec{A}_t)\,s $.
Assume we have observed $\vec{o}_{0}=\{\vec{a}_{0}, \vec{y}_{0}\}$ and $\vec{y}_t$. To apply the CFD in Theorem~\ref{thm:CFD} we make the following identifications
\begin{align*}
	A\leftarrowtail \vec{A}_{t},& 
	\quad Y\leftarrowtail \vec{Y}_{t+s}, 
	\quad  M \leftarrowtail \vec{Z}_{t+s}, 
	\quad W \leftarrowtail \{\vec{Z}_t,\vec{Y}_t, \vec{A}_{0},\vec{Y}_{0}\}=\{\vec{Z}_t,\vec{Y}_t, \vec{O}_{0}\}\,.
\end{align*}
Although $\vec Z_t$ is latent, observability in this discrete-time representation implies that it can be uniquely reconstructed from the treatment-outcome history. By Assumption~\ref{ass:latent}, the mediator distribution under treatment $\vec a_t$ is then given by $p(\vec z_{t+s}|\vec a_t,\vec z_t)$. Consequently, both the conditioning state and the mediator distribution required by the CFD adjustment are identifiable.

We can check graphically in Fig.~\ref{fig:fig1dags}(a,b) that $M=\vec{Z}_{t+s}$ satisfies the CFD criterion relative to the ordered pair $(A,Y)=(\vec{A}_t, \vec{Y}_{t+s})$, if we use $W=\{\vec{Z}_t,\vec{Y}_t, \vec{O}_{0}\}$ as conditioning set. Now we compute the probability density for the potential outcome $\vec{Y}_{t+s}(\vec{a}_t)$ when applying treatment $\vec{A}_t=\vec{a}_t$ at time $t$ using \eqref{eq:cfd_adjust} and replacing the sums by integrals
\begin{align} \label{eq:adjust_full_hidden_confounder}
	P_{\vec{Y}_{t+s}(\vec{a}_t)}\left(\vec{y}_{t+s}\right) &= \int_{\vec{z}_{t+s}}\int_{\vec{z}_{t},\vec{y}_t,\vec{o}_0} \int_{\vec{a}'_t}
	\left\{ p\left(\vec{y}_{t+s}|\vec{a}'_t,\vec{z}_{t+s},\vec{z}_{t},\vec{y}_t,\vec{o}_0\right)
	\right.\\
	&\qquad \cdot \left. p\left(\vec{a}'_t|\vec{z}_{t},\vec{y}_t,\vec{o}_0\right)\,
	p\left(\vec{z}_{t+s}|\vec{a}_t,\vec{z}_{t},\vec{y}_t,\vec{o}_0\right)\,
	p\left(\vec{z}_{t},\vec{y}_t,\vec{o}_0\right)\right\}.
\end{align}
By Assumptions~\ref{ass:observation} and \ref{ass:latent} we have the following independence relations
\begin{align*}
	p\left(\vec{y}_{t+s}|\vec{a}'_t,\vec{z}_{t+s},\vec{z}_{t},\vec{y}_t,\vec{o}_0\right) &= p\left(\vec{y}_{t+s}|\vec{z}_{t+s}\right)\\
	p\left(\vec{z}_{t+s}|\vec{a}_t,\vec{z}_{t},\vec{y}_t,\vec{o}_0\right) &= p\left(\vec{z}_{t+s}|\vec{a}_t,\vec{z}_{t}\right).
\end{align*}
Integrating over $\vec{a}'_t$ and using $p\left(\vec{z}_{t},\vec{y}_t,\vec{o}_0\right)=p\left(\vec{z}_{t}|\vec{y}_t,\vec{o}_0\right)\,p\left(\vec{y}_t,\vec{o}_0\right)$ we find
\begin{align*}
	P_{\vec{Y}_{t+s}(\vec{a}_t)}\left(\vec{y}_{t+s}\right) &= \int_{\vec{z}_{t+s}}\int_{\vec{z}_{t},\vec{y}_t,\vec{o}_0} 
	\left\{ p\left(\vec{y}_{t+s}|\vec{z}_{t+s}\right) 
	p\left(\vec{z}_{t+s}|\vec{a}_t,\vec{z}_{t}\right)\,
	p\left(\vec{z}_{t}|\vec{y}_t,\vec{o}_0\right)p\left(\vec{y}_t,\vec{o}_0\right)\right\},
\end{align*}
from which we read the conditional density for the potential outcome $\vec{Y}_{t+s}(\vec{a}_t)$ when applying treatment $\vec{A}_t=\vec{a}_t$ at time $t$, given we have observed outcome $\vec{y}_t$ at time $t$ and the treatment-outcome history $\vec{o}_{0}$ at time zero
\begin{align*}
	P_{\vec{Y}_{t+s}(\vec{a}_t)|\vec{Y}_t, \vec{O}_{0}}\left(\vec{y}_{t+s}\left|\vec{y}_t, \vec{o}_{0}\right.\right)	= \int_{\vec{z}_{t+s}}\int_{\vec{z}_{t}} 
	\left\{ p\left(\vec{y}_{t+s}|\vec{z}_{t+s}\right) 
	p\left(\vec{z}_{t+s}|\vec{a}_t,\vec{z}_{t}\right)\,
	p\left(\vec{z}_{t}|\vec{y}_t,\vec{o}_0\right)\right\}.
\end{align*}
The adjustment formula remains valid for dynamic treatment regimes in which the current observation $\vec{y}_t$ and the observed history $\vec{o}_0$ determine the treatment $\vec{a}_t$, because the same CFD criterion continues to hold (Fig.~\ref{fig:fig1dags}(b)).
\end{proof}

The key requirement is that the latent state is observable from the observed history.  For the deterministic ODE in \eqref{eq:node_ssm} the dynamics is given by the point mass: $p(\vec{z}_{t+s}|\vec{a}_{[t,t+s)},\vec{z}_t)=\delta(\vec{z}_{t+s}-\vec{z}_t-\int_{t}^{t+s}\vec{f}(\vec{z}_\tau,\vec{a}_\tau)\,d\tau)$. If we replace the ODE by a stochastic differential equation we need a notion of stochastic observability \citep{van_handel_observability_2009}.

\subsection{Neural ODEs in observable normal form (ObsNODEs)}
The identification formula \eqref{eq:outcome_ident} requires that the latent state can be inferred from observed outcomes and treatment history. Standard latent Neural ODEs do not guarantee this property. We therefore parameterize the latent dynamics in continuous triangular observable normal form \citep{bernard_observer_2019,adamy_nonlinear_2024}:
\begin{alignat}{3}
	\dot{\vec{Z}}_t =
	\begin{pmatrix}
		\vec{\dot{Z}}_t^{(1)}\\
		\vdots\\
		\vec{\dot{Z}}_t^{(i)}\\
		\vdots\\
		\vec{\dot{Z}}_t^{(m)}
	\end{pmatrix}
	&=
	\begin{pmatrix}
		\vec{Z}_t^{(2)} + \vec{\phi}^{(1)}_{\theta_1}(\vec{Z}_t^{(1)},\vec{a}_t)\\
		\vdots\\
		\vec{Z}_t^{(i+1)} + \vec{\phi}^{(i)}_{\theta_i}(\vec{Z}_t^{(1)},\ldots,\vec{Z}_t^{(i)},\vec{a}_t)\\
		\vdots\\
		\vec{\phi}^{(m)}_{\theta_m}(\vec{Z}_t,\vec{a}_t)
	\end{pmatrix},
	\qquad
	\vec{Y}_t = \vec{Z}_t^{(1)}.
	\label{eq:triang_nf}
\end{alignat}
Here the state  $\vec{Z}_t = \left(\vec{Z}_t^{(1)},\ldots, \vec{Z}_t^{(m)}\right)\in \mathbb{R}^{d_z}$ is partitioned into $m$ blocks $\vec{Z}_t^{(i)}\in\mathbb{R}^{d_y}$, so that $d_z=md_y$, and each $\vec{\phi}^{(i)}_{\theta_i}$ is implemented by a neural network. Under suitable conditions, observable systems can be transformed into this normal form without loss of expressiveness~\citep{bernard_observer_2019}. 

We allow for additive observation noise,
$\vec{Y}_t=\vec{Z}_t^{(1)}+\vec{\eta}_t$,  where $\vec{\eta}_t$ is a centered noise process. 
This noise does not affect identification. The hidden disturbance ($\vec{\epsilon}_t$) need not be centered and is accounted for by the corresponding adjustment formula.

\subsection{Prediction and training of ObsNODEs}

We approximate the filtering distribution $p(\vec{z}_t | \vec{y}_t,\vec{o}_{[0,t)})$ with an RNN encoder, following \citep{chen_neural_2018}. Given the inferred state $\vec{z}_t$, forecasts under a hypothetical treatment path $\vec{a}_{[t,t+s)}$ are obtained by integrating \eqref{eq:triang_nf} forward in time and projecting to the observed space.

Training is self-supervised. For each unit, we choose a decision time $t_c$ and provide observations up to $t_c$. The model then predicts future observations on $(t_c,t_f]$. 

Observations are available only at irregular time points $\Tau_{ij}$.
Let $\Tau_{ij}(t_c,t_f)=\Tau_{ij}\cap (t_c,t_f]$ denote the observation times of component $j$ for unit $i$ in that forecast window $(t_c,t_f]$. The predicted outcome vector for unit $i$ and outcome component $j$ at time $t$ is to be compared against  the corresponding observed value $y_{tij}^{dat}$ in the training data. 
We minimize the masked squared error 
\begin{align*}
	\mathcal{L}_{(t_c,t_f]} =
	\sum_{i=1}^{n}\sum_{j=1}^{d_y}
	\frac{1}{n\,\sigma_j^2\,|\Tau_{ij}(t_c,t_f)|}
	\sum_{t\in\Tau_{ij}(t_c,t_f)}
	\left(y_{tij}-y_{tij}^{\mathrm{dat}}\right)^2,
\end{align*}
where $|\Tau_{ij}(t_c,t_f)|$ denotes the number of observations of component $j$ for unit $i$ in the forecasting interval $(t_c,t_f]$. This normalization compensates for varying observation frequencies across units and outcome components. The scaling factor $\sigma_j^2$ is the empirical variance of the 
$j-$th outcome component across all units and observed time points in the training data and can be set to 1 for standardized data. 

During training, a decision time $t_c\in [0,t_f)$ is selected for all units in a batch of size $n$ and the corresponding losses $\mathcal{L}_{(t_c,t_f]}$ are averaged over different batches. We found that both a fixed set of possible values for $t_c$ evenly distributed over $[0,t_f)$ or a uniform random choice led to very similar results. Choosing different decision times $t_c$ in each batch exposes the model to varying assimilation times and forecast horizons. 

\section{Experiments}

We evaluate ObsNODE on synthetic and semi-synthetic datasets where the ground-truth causal effect of a prescribed treatment path on the outcome trajectories is known and, in addition,  on a real world data set with an external validation. We compare against the IGC-Net \citep{hess2026igcnet}, SCIP-Net \citep{hess2025scipnet} and two Treatment Effect Neural Controlled Differential Equation models \citep{seedat_continuous-time_2022}: optAB for categorical treatments \citep{wendland_optimal_2024} and doseAI for categorical and continuous treatments \citep{bellazzi_dynamic_2025}. 

Forecast root mean squared error (RMSE) was scaled with the standard deviation of each dynamic variable in the test set and is reported as a function of assimilation time (the duration of the treatment-observation history $t_c$) and prediction horizon (Figures~\ref{fig:res_cancer}--\ref{fig:resmimic4}). The RMSE values of these heatmaps are given in \url{https://github.com/JenniferJaschob/ObsNODE.git}.  

As an ablation study, we compare ObsNODE with an otherwise identical latent Neural ODE without the observable normal form. For each test patient, both models assimilate the factual history up to $t_c=5$ and predict outcomes under five prespecified treatment regimes. Counterfactual bias (CB) is the signed difference between predicted and ground-truth potential outcomes, averaged over all test patients and six prediction time points. We also evaluate the average treatment effect (ATE) of each regime relative to a baseline treatment (Tables~\ref{tab:cb_cancer} and~\ref{tab:cb_semisynmimiciv}).

Results are averaged over five independent runs with disjoint training, validation, and test sets.  For synthetic and semi-synthetic data we vary the confounding strength $\gamma$. Details on data generation and the counterfactual treatment scenarios are provided in the Appendix \ref{subsec: tumor_growth_simulator} and  \ref{subsec:semi_synthetic_data_simulation}.

\subsection{Experiments with fully-synthetic data}

\textbf{Data:} The synthetic dataset is generated from a dynamic model of tumor volume \citep{geng_prediction_2017} and body weight \citep{bellazzi_dynamic_2025} under chemotherapy and radiotherapy. The treatment vector $\vec{a}_t$ represents chemo- and radiotherapy doses administered in monthly cycles, and the outcome $\vec{y}_t$ consists of tumor volume and body weight.

To simulate time-dependent confounding, treatment probabilities depend on the mean tumor volume over the previous 15 days~\citep{melnychuk_causal_2022}. Patient heterogeneity is introduced by sampling tumor growth and treatment response parameters from predefined distributions (Appendix~\ref{subsec: tumor_growth_simulator}).

\begin{figure}[htb]
	%\vskip 0.2in
	\begin{center}
		\centerline{\includegraphics[width=\textwidth]{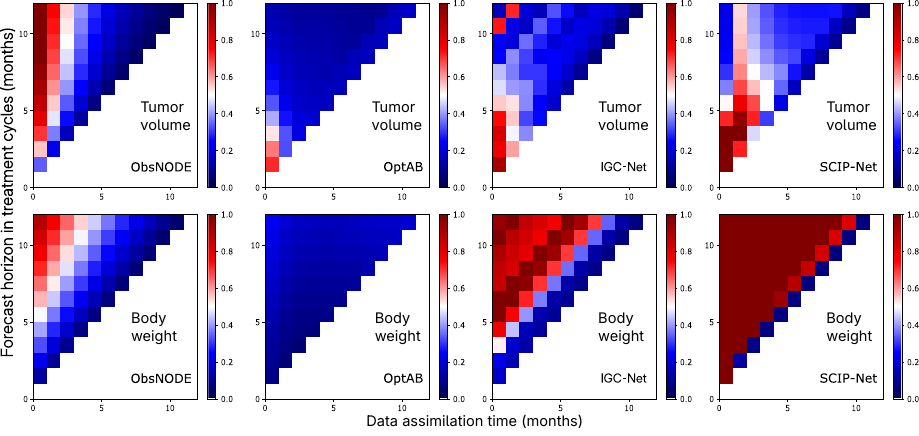}}
		\caption{\textbf{Synthetic cancer dataset.} RMSE heatmaps (mean over five runs) for standardized (Z-score) tumor volume and body weight forecasts at confounding strength $\gamma=4$. The horizontal axis shows the assimilation time and the vertical axis the forecast horizon. Columns correspond to ObsNODE, doseAI~\citep{bellazzi_dynamic_2025}, IGC-Net~\citep{hess2026igcnet} and SCIP-Net~\citep{hess2025scipnet}. RMSE values larger than one were clipped at one.}
		\label{fig:res_cancer}
	\end{center}
	\vskip -0.2in
\end{figure}
\begin{table*}[htb]
	%\vskip 0.15in
	\begin{center}
		\begin{small}
			\begin{sc}
				\begin{tabular}{llcccccc}
					\hline
					%\multicolumn{7}{c}{Tumor volume}\\
					%\hline
					$\gamma$ & {Model/ $s$} & 1&2&3&4&5&6 \\ 
					\hline
					%\multicolumn{7}{c}{$\gamma=4$}\\
					%\hline
					4 & ObsNODE  & \textbf{0.06±0.00}  &    \textbf{0.11±0.00} &    0.14±0.00 & 0.17±0.00 & 0.19±0.00 &   0.21±0.01  \\
					& DoseAI  &  0.16±0.02 & 0.15±0.02 & \textbf{0.12±0.01} &     \textbf{0.12±0.01} &  \textbf{0.11±0.00} &   \textbf{0.10±0.01}  \\
					& IGC-Net  &  0.13±0.01  &  0.18±0.04 & 0.22±0.05  & 0.22±0.03 & 0.17±0.01 & 0.26±0.14  \\
					& SCIP-Net & 0.16±0.00 & 0.27±0.02 & 0.32±0.04  & 0.30±0.05 & 0.28±0.06 & 0.27±0.07\\
                    & NODE & 0.61±0.00 & 0.47±0.00 & 0.37±0.00 & 0.30±0.00 & 0.25±0.00 & 0.22±0.00\\
					\hline
					%\multicolumn{7}{c}{$\gamma=6$}\\
					%\hline
					6 & ObsNODE & \textbf{0.06±0.00}  &    \textbf{0.10±0.00} &    0.13±0.00 & 0.15±0.00 & 0.17±0.00 &   0.18±0.01  \\
					& DoseAI  &  0.12±0.02 & 0.11±0.01 & \textbf{0.10±0.01} &     \textbf{0.10±0.01} &  \textbf{0.10±0.00} &   \textbf{0.10±0.01}  \\
					& IGC-Net  &  0.12±0.01  &  0.18±0.03 & 0.20±0.04  & 0.20±0.03 & 0.16±0.01 & 0.25±0.15 \\
					& SCIP-Net & 0.14±0.00 & 0.24±0.02 & 0.31±0.05 & 0.29±0.05 & 0.27±0.06 & 0.26±0.06\\
                    & NODE & 0.56±0.00 & 0.43±0.00 & 0.34±0.00 & 0.28±0.00 & 0.24±0.00 & 0.20±0.00\\
					\hline
					%\multicolumn{7}{c}{$\gamma=8$}\\
					%\hline
					8 & ObsNODE & \textbf{0.05±0.00}  &    \textbf{0.09±0.00} &    0.12±0.00 & 0.14±0.00 & 0.16±0.00 &   0.18±0.01  \\
					& DoseAI  &  0.12±0.01 & 0.11±0.00 & \textbf{0.11±0.00} &     \textbf{0.11±0.01} &  \textbf{0.12±0.01} &   \textbf{0.13±0.02}  \\
					& IGC-Net &  0.11±0.01  &  0.18±0.03 & 0.20±0.04  & 0.19±0.03 & 0.16±0.01 & 0.25±0.17 \\
					& SCIP-Net & 0.14±0.00 & 0.24±0.02 & 0.31±0.04 & 0.29±0.05 & 0.28±0.06 & 0.27±0.06\\
                    & NODE & 0.55±0.00 & 0.43±0.00 & 0.35±0.00 & 0.29±0.00 & 0.25±0.00 & 0.22±0.00\\
					\hline
				\end{tabular}
			\end{sc}
			\caption{\textbf{The RMSE for the tumor volume from synthetic data.}  The RMSE  between predicted and observed tumor volume (mean ± std over five runs) for a fixed data assimilation time of $t_c=5$  and different forecast horizons $s = t_f - t_c$ (in monthly treatment cycles) at different levels of confounding~$\gamma$.}
			\label{tab:add_res_tumorvolumen}
		\end{small}
	\end{center}
	\vskip -0.1in
\end{table*}
\begin{table}[htb]
%\vskip 0.15in
	\begin{center}
	\begin{small}
	\begin{sc}
    \begin{tabular}{lclccccc}
    \hline
        Metric & Variable & Model  / $CF_k$ & 1 & 2 & 3 & 4 & 5 \\
    \hline

        CB & Tumor Volume &  ObsNODE &  \textbf{0.1738} & \textbf{0.1739} & \textbf{0.1724} & \textbf{0.1708} & \textbf{0.1708} \\
       &  & NODE   &  -0.5152 & -0.5288 & -0.5364 & -0.5321 & -0.5321\\

    \hline

    CB &  Body Weight & ObsNODE &  \textbf{0.0952} & \textbf{0.0941} & \textbf{0.0924} & \textbf{0.0924} & \textbf{0.0926}  \\
    &  & NODE  & -2.0229 & -2.0272 & -2.0313 & -2.0315 & -2.0315 \\
    
    \hline
    \hline   
    ATE &    Tumor Volume & True & -0.0105 & 0.0031 & 0.0107 & 0.0063 & 0.0063 \\
        & & ObsNODE &  \textbf{0.1633} & \textbf{0.1770} & \textbf{0.1831} & \textbf{0.1771} & \textbf{0.1771} \\
       &  & NODE & -0.5257 & -0.5257 & -0.5257 & -0.5257 & -0.5257 \\
        
    \hline     
    ATE &    Body Weight & True & -0.0016 & 0.0227 & 0.0068 & 0.0070 & 0.0070 \\
     
     &   & ObsNODE & \textbf{0.0937} & \textbf{0.0968} & \textbf{0.0992} & \textbf{0.0995} & \textbf{0.0995} \\
      &   & NODE & -2.0245 & -2.0245 & -2.0245 & -2.0245 & -2.0245 \\
         \hline
    \end{tabular}
    \end{sc}
    \caption{
    \textbf{Counterfactual bias (CB) and estimated average treatment effect (ATE)} relative to the baseline treatment for five treatment regimes ($CF_1$--$CF_5$) in the synthetic cancer dataset, with data assimilation time $t_c=5$ and confounding level $\gamma=4$. CB  and ATE were averaged over six prediction time points, and all results are averaged over five runs.}
    \label{tab:cb_cancer}
    \end{small}
	\end{center}
    \vskip -0.2in
\end{table}

\textbf{Results:} Figure~\ref{fig:res_cancer} compares tumor volume and body weight forecast RMSE for $\gamma=4$. For long horizons, doseAI achieves the lowest error, followed by ObsNODE and IGC-Net, whereas ObsNODE performs best for short horizons ($\tau \leq 2$ cycles). Detailed RMSE values for the tumor volume at different confounding strengths $\gamma$ and prediction horizons $\tau$ are reported in Table~\ref{tab:add_res_tumorvolumen}. ObsNODE consistently shows the smallest variance across different runs.

ObsNODE consistently outperforms the Neural ODE (NODE) in CB and ATE estimation (Table~\ref{tab:cb_cancer}), reducing the mean absolute ATE error from 0.5289 to 0.1723 for tumor volume and from 2.0289 to 0.0933 for body weight.

\subsection{Experiments with semi-synthetic data}
\textbf{Data:} The second experiment employs a semi-synthetic dataset derived from real-world medical data. The MIMIC-IV dataset \citep{johnson_mimic-iv_2023} serves as the foundation, with a focus on the patient cohort diagnosed with sepsis. Based on this patient cohort, synthetic data are generated to represent patient trajectories and outcomes under both endogenous and exogenous dependencies, while explicitly incorporating treatment effects. The synthetic data generation methodology was originally defined using the MIMIC-III dataset \citep{schulam_reliable_2018, johnson_mimic-iii_2016,melnychuk_causal_2022}. 
Further details on the synthetic data generation process are provided in Appendix \ref{subsec:semi_synthetic_data_simulation}.
\begin{figure}[htb]
	%\vskip 0.2in
	\begin{center}
		\centerline{\includegraphics[width=\columnwidth]{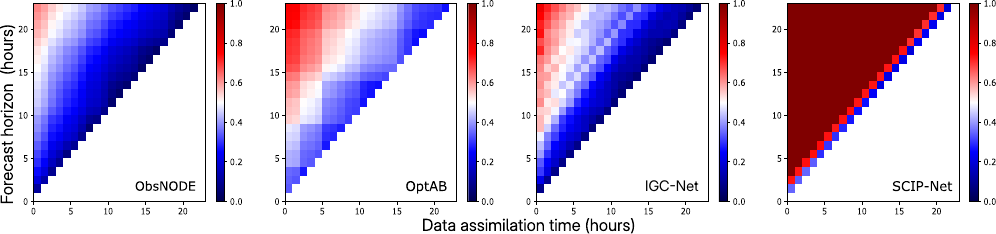}}
        \caption{\textbf{Semi-synthetic MIMIC-IV sepsis dataset.} RMSE for ObsNODE, OptAB, IGC-Net and SCIP-Net as a function of data assimilation time (observation time) and forecast horizon over a 24-hour horizon. Results are averaged over five runs and RMSE values larger than one were clipped at one.}
		\label{fig:res_semisynmimic4}
	\end{center}
	\vskip -0.3in
\end{figure}
\begin{table*}[htb]
	%\vskip 0.15in
	\begin{center}
		\begin{small}
			\begin{sc}
				\begin{tabular}{lcccccc}
					\hline
					{Model/ $s=t_f-t_c$} & 1&2&3&4&5&6 \\
					%\hline
					%\multicolumn{7}{c}{\textbf{Semi-synthetic outcome}}\\
					\hline
					ObsNODE & 0.13±0.00 & \textbf{0.21±0.00}  & \textbf{0.25±0.00}  & \textbf{0.27±0.00}  & \textbf{0.30±0.00}  & \textbf{0.33±0.00}   \\
					OptAB  & 0.34±0.04 & 0.35±0.04  & 0.41±0.02 & 0.42±0.02 & 0.44±0.02  & 0.45±0.02   \\
					IGC-Net &\textbf{0.10±0.03} & \textbf{0.21±0.01}  & \textbf{0.25±0.01} & 0.29±0.02  & 0.34±0.02  & 0.38±0.01 \\
					SCIP-Net & 0.34±0.10 & 0.73±0.10 & 1.32±0.16 & 1.32±0.16  & 1.33±0.16  & 1.32±0.16 \\
                    
                    NODE & 0.88±0.06 & 0.63±0.11 & 0.48±0.12 & 0.43±0.06 & 0.41±0.04 & 0.42±0.05\\
					\hline
				\end{tabular}
			\end{sc}
			\caption{\textbf{Semi-synthetic MIMIC-IV sepsis dataset.} RMSE (mean $\pm$ std over 5 runs) over a $s=t_f-t_c$ hour horizon when observing for a fixed data assimilation time of $t_c=1$ hour.}
			\label{tab:add_res_semisynmimiciv}
		\end{small}
	\end{center}
	\vskip -0.1in
\end{table*}
\begin{table}[htb]
%\vskip 0.15in
    \begin{center}
	\begin{small}
	\begin{sc}
    \begin{tabular}{lclccccc}
    \hline
     Metric &   Variable & Model / $CF_k$ & 1 & 2 & 3 & 4 & 5 \\
    \hline   
     CB &    Semi-synthetic &ObsNODE & \textbf{1.9165} & \textbf{1.2978} & \textbf{1.3962} &\textbf{2.6466} & \textbf{-0.0149}  \\
      &   outcome & NODE & 5.1987 & 2.4121 & 2.4121 &  7.6795 & -0.5663\\
     \hline
        \hline
     ATE &    Semi-synthetic  & True & -5.7596 & -5.1849 & -5.1849 & -5.1849 & -3.9458 \\
        & outcome & ObsNODE & \textbf{-3.8431} &  \textbf{-3.8860} & \textbf{-3.7887} & \textbf{-3.9403} & \textbf{-3.7794} \\
       &  & NODE & -0.5608 & -2.7728 & -2.7728 & 1.2556 & -4.5121 \\
    \hline
    \end{tabular}
    \end{sc}
    \caption{\textbf{Counterfactual bias (CB) and estimated average treatment effect (ATE)}  relative to the baseline treatment for five treatment regimes  ($CF_1$--$CF_5$) in the semi-synthetic MIMIC-IV sepsis dataset, with data assimilation time $t_c=5$. CB and ATE were averaged over six prediction time points, and all results are averaged over five runs.}
    \label{tab:cb_semisynmimiciv}
    \end{small}
	\end{center}
%\vskip -0.1in    
\end{table}

\textbf{Results:} Figure~\ref{fig:res_semisynmimic4} shows results on the semi-synthetic MIMIC-IV sepsis dataset. ObsNODE achieves consistently a low RMSE across the 24-hour prediction horizon and performs better than OptAB, IGC-Net and SCIP-Net for longer horizons. RMSE statistics averaged over five runs (Table~\ref{tab:add_res_semisynmimiciv}
) show very small variance for ObsNODE, indicating stable predictions compared to the baselines.

For the semi-synthetic data, ObsNODE consistently achieves lower CB across all counterfactual treatment scenarios and reduces the mean absolute ATE error from 3.6537 to 1.4523 compared with the standard Neural ODE (NODE) without the observable normal form (Table~\ref{tab:cb_semisynmimiciv}).

\subsection{Experiments with real-world data}
\textbf{Data:} Selecting antibiotic therapy for sepsis patients is challenging~\citep{wendland_optimal_2024}. We use the Sequential Organ Failure Assessment (SOFA)  score, a summary measure of organ dysfunction, as the primary outcome and predict the effects of the antibiotics vancomycin, ceftriaxone, and piperacillin-tazobactam on the SOFA trajectory. In addition, we predict treatment effects on the side-effect markers creatinine, total bilirubin, and alanine aminotransferase (ALT). Further dynamic variables included in the observation vector are listed in Table~\ref{tab:time_dependent_covariates_mimiciv} in Appendix~\ref{subsec:mimicIV_data}.

%The MIMIC-IV sepsis cohort~\citep{johnson_mimic-iv_2023} was used for training and testing, while ObsNODE was externally validated on the independent AmsterdamUMCdb cohort~\citep{thoral_sharing_2021}.
The MIMIC-IV sepsis cohort~\citep{johnson_mimic-iv_2023} was used to train and test all models. In addition, ObsNODE was externally validated on the independent AmsterdamUMCdb sepsis cohort~\citep{thoral_sharing_2021}.

\begin{figure*}[htb]
\centering
\includegraphics[width=0.95\textwidth]{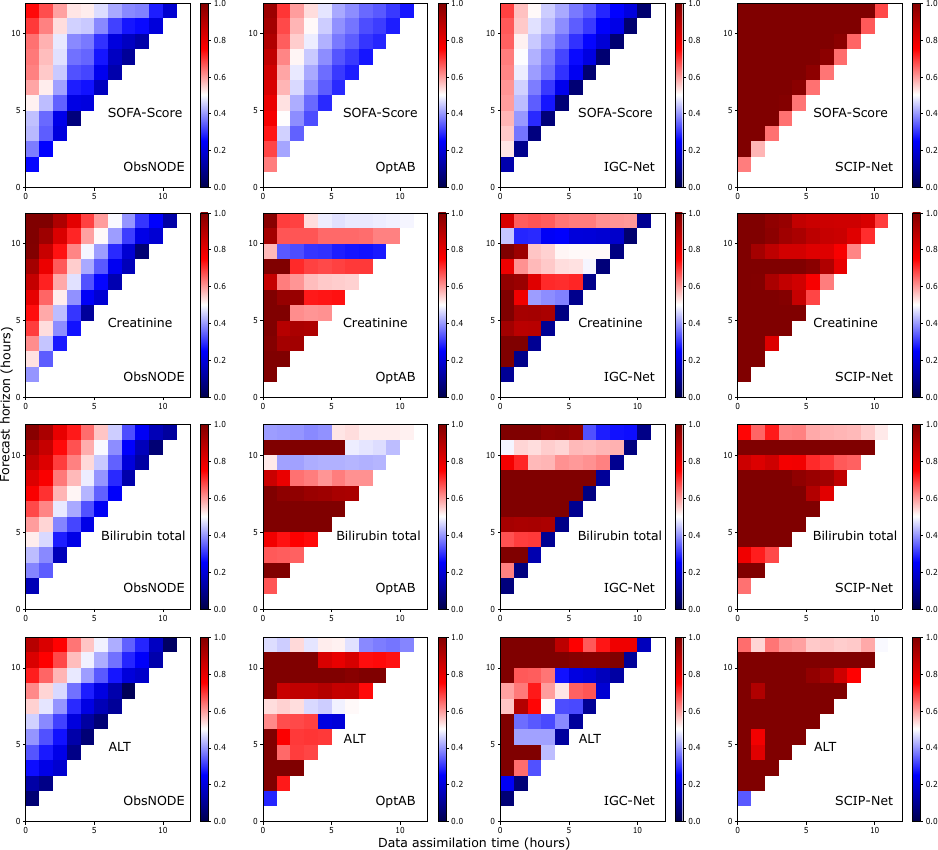}
\caption{Forecasting performance on MIMIC-IV. Forecast RMSE for SOFA score, creatinine, total bilirubin, and alanine aminotransferase (ALT) across data assimilation times and forecast horizons.}
\label{fig:resmimic4}
\end{figure*}

\textbf{Results:} 
Table~\ref{tab:add_res_mimiciv} summarizes the five-run robustness analysis. ObsNODE shows low variability across datasets sampled from the same distribution. Its competitive performance is particularly evident at longer forecast horizons, while IGC-Net performs better for very short-term predictions of selected laboratory variables (Figure~\ref{fig:resmimic4}).

Table~\ref{tab:add_res_amsterdam} summarizes the external validation of ObsNODE, trained on MIMIC-IV, on AmsterdamUMCdb. Predictions were more accurate for creatinine and total bilirubin and remained comparable for SOFA score and ALT. The improved performance may reflect the higher observation density in AmsterdamUMCdb, which provides more measurements for data assimilation during prediction.
\begin{table*}[htb]
    \centering
    \begin{subtable}{\textwidth}
        \centering
        \begin{small}
        \begin{sc}
        \begin{tabular}{lcccccc}
            					\hline
					{Model/ $s=t_f-t_c$} & 1&2&3&4&5&6 \\
					\hline
					\multicolumn{7}{c}{\textbf{Sofa-Score}}\\
					\hline
					ObsNODE & 0.24±0.00 & \textbf{0.32±0.01} & \textbf{0.37±0.01} & \textbf{0.42±0.02} & 0.47±0.03 & \textbf{0.50±0.04}  \\
					OptAB  & 0.41±0.01 & 0.46±0.01 & 0.50±0.02 & 0.53±0.02 & 0.57±0.02 & 0.59±0.02  \\
					IGC-Net & \textbf{0.09±0.00} & 0.37±0.00 & 0.40±0.00 & 0.43±0.00 & \textbf{0.45±0.00} & 0.48±0.01 \\
					SCIP-Net & 0.57±0.06 & 1.03±0.44 & 1.25±0.43 & 1.23±0.39 & 1.21±0.35 & 1.19±0.33 \\
					\hline
					\multicolumn{7}{c}{\textbf{Creatinine}}\\
					\hline
					ObsNODE & 0.34±0.00 & \textbf{0.44±0.00} & \textbf{0.53±0.00} & \textbf{0.53±0.02} & \textbf{0.58±0.04} & 0.65±0.07 \\
					OptAB & 1.14±0.02 & 0.93±0.02 & 0.89±0.03 & 1.23±0.03 & 0.96±0.01 & \textbf{0.62±0.02} \\
					IGC-Net & \textbf{0.09±0.02} & 1.13±0.01 & 0.89±0.00 & 0.92±0.01 & 0.78±0.15 & 0.95±0.01 \\
                    SCIP-Net  & 1.01±0.04  & 1.06±0.26 & 1.36±0.28  &  1.39±0.30 & 1.17±0.39 &  0.99±0.42\\
					
					\hline
					\multicolumn{7}{c}{\textbf{Bilirubin-total}}\\
					\hline
					ObsNODE & 0.34±0.00 & \textbf{0.38±0.00} & \textbf{0.46±0.02} & \textbf{0.54±0.04} & \textbf{0.62±0.07} & \textbf{0.70±0.11} \\
					OptAB  & 1.36±0.05 & 0.66±0.01 & 0.73±0.07 & 1.05±0.04 & 1.56±0.14 & 0.97±0.05 \\
					IGC-Net  & \textbf{0.07±0.01} & 1.40±0.02  & 0.69±0.01 & 0.91±0.00 & 1.07±0.01 & 1.59±0.03 \\
					
                    SCIP-Net & 1.23±0.17 & 0.72±0.12 & 1.13±0.11 & 1.24±0.12 & 1.84±0.14 & 1.17±0.14 \\
					\hline
					\multicolumn{7}{c}{\textbf{ALT}}\\
					\hline
					ObsNODE & 0.09±0.00 & \textbf{0.22±0.01} & \textbf{0.27±0.03} & \textbf{0.33±0.06} & \textbf{0.39±0.10} & \textbf{0.47±0.13} \\
					OptAB  & 0.72±0.20 & 1.25±0.09 & 0.65±0.16 & 0.76±0.25 & 0.67±0.21 & 0.54±0.11 \\
					IGC-Net  & \textbf{0.06±0.02} & 0.64±0.17  & 1.34±0.21 & 0.41±0.02  & 0.36±0.03 & 0.90±0.23 \\
                    SCIP-Net & 1.75±0.20 & 1.51±0.30 & 0.81±0.28 & 0.77±0.37 & 2.24±0.45 & 1.45 ± 0.26 \\
					\hline
        \end{tabular}
        \end{sc}
        \end{small}
        \caption{RMSE for ObsNODE, OptAB, IGC-Net and SCIP-Net on a test set from the MIMIC-IV sepsis dataset.}
        \label{tab:add_res_mimiciv}
    \end{subtable}
    \begin{subtable}{\textwidth}
        \centering
        \begin{small}
        \begin{sc}
        \begin{tabular}{lcccccc}
            					\hline
					Variable/ $s=t_f-t_c$ & 1&2&3&4&5&6 \\
					\hline
					Sofa-Score & 0.26±0.00 & 0.37±0.01 & 0.43±0.02 & 0.44±0.02  & 0.51±0.02 & 0.59±0.03  \\
					Creatinine & \textbf{0.14±0.00} & \textbf{0.18±0.02} & \textbf{0.22±0.05} & \textbf{0.27±0.09}  & \textbf{0.34±0.13} & \textbf{0.39±0.17}  \\
					Bilirubin-total & \textbf{0.01±0.00}   & \textbf{0.11±0.02} & \textbf{0.18±0.05} & \textbf{0.25±0.09} & \textbf{0.33±0.13} & \textbf{0.40±0.17} \\
					ALT & \textbf{0.01±0.00} & 0.28±0.01 & 0.34±0.03 & 0.41±0.05 & 0.50±0.07 & 0.86±0.07 \\
					\hline
        \end{tabular}
        \end{sc}
        \end{small}
        %\caption{RMSE for ObsNODE trained on MIMIC-IV and tested on the external AmsterdamUMCdb validation set. Bold values indicate lower RMSE on AmsterdamUMCdb than on the MIMIC-IV test set reported above.}
        \caption{RMSE for ObsNODE, trained on MIMIC-IV sepsis data and tested on an external validation set from AmsterdamUMCdb. Bold values indicate lower RMSE on AmsterdamUMCdb than on the MIMIC-IV test set reported above.}
        \label{tab:add_res_amsterdam}
    \end{subtable}
    \caption{\textbf{Forecasting results on the real-world Sepsis dataset.}
    RMSE for different prediction horizons (in hours) and a fixed data assimilation time of $t_c=1$ hour. Mean and standard deviation are
    reported across five runs.}
    \label{tab:add_res_sepsis}
    \vskip -0.2in
\end{table*}

\section{Related Work}
Classical statistical approaches for longitudinal causal inference focus on estimating population-level average potential outcomes and do not condition on individual treatment and outcome histories~\citep{robins_new_1986,bang_doubly_2005,lok_statistical_2008,van_der_laan_targeted_2012,rytgaard_continuous-time_2022}. Consequently, they cannot be used for individualized outcome forecasting. These approaches are typically grounded in structural causal models and identification results such as the longitudinal g-formula. Extensions to continuous time include structural dynamic causal models~\citep{bongers_causal_2022}, which adopt a different formulation of interventions than considered here. Recent work has developed discrete-time sequence models for estimating conditional average potential outcomes (CAPO) (i.e. the expectation value of \eqref{eq:outcome_ident}), including recurrent and attention-based architectures
\citep{bica_estimating_2020, li_g-net_2020, melnychuk_causal_2022, hess2026igcnet}. 
A limitation of these approaches is their reliance on fixed time discretizations, whereas clinical measurements are typically observed at irregular time points.

Continuous-time approaches based on neural differential equations address this limitation. Treatment Effect Controlled Differential Equations (TE-CDEs)~\citep{seedat_continuous-time_2022} combine controlled differential equations~\citep{kidger_neural_2020} with representation balancing to mitigate time-dependent confounding. The original formulation considered binary treatments and was later extended to continuous treatments~\citep{bellazzi_dynamic_2025}. However, balancing-based approaches may not fully address time-dependent confounding in sequential settings~\citep{melnychuk_causal_2022}. 

The SCIP-Net~\citep{hess2025scipnet} is a recent continuous time method for estimating the CAPO based on a stabilized inverse propensity score weighting, which is, however, restricted to discrete treatment levels.

Our ObsNODE framework is most closely related to CF-ODE~\citep{brouwer_predicting_2022}, which parameterizes a continuous-time state-space model of the form~\eqref{eq:node_ssm} using a Neural ODE. CF-ODE assumes that observations at treatment time uniquely determine the latent state, corresponding to the observability condition (Definition~\ref{def:observability}). However, this assumption is not structurally enforced and is difficult to verify in practice. In contrast, ObsNODE employs the observable normal form~\eqref{eq:triang_nf}, which guarantees observability by construction and makes explicit the link between control-theoretic observability and the identifiability condition required for the causal adjustment formula~\eqref{eq:outcome_ident}. Moreover, CF-ODE conditions only on prior observations and does not explicitly incorporate the treatment history.

Together, these limitations motivate a continuous-time state-space formulation that enforces observability and enables causal outcome prediction through an explicit adjustment formula, which we develop in the ObsNODE framework.

\section{Conclusion and Limitations}
We introduced Observable Neural ODEs (ObsNODEs), linking control-theoretic observability to causal identifiability in continuous-time state-space models. ObsNODEs enforce observability by construction, enable causal forecasting under alternative treatment trajectories, and perform strongly on synthetic, semi-synthetic, and clinical data, including external validation.

The approach relies on consistency, positivity, and the stated state-space assumptions. The tractable adjustment formula permits contemporaneous hidden confounding but assumes temporally independent disturbances; serially correlated confounding requires the general path-space adjustment derived in the Appendix. State reconstruction from finite, noisy observations remains approximate, and causal accuracy cannot be assessed on clinical data without counterfactual ground truth. Future work should address persistent confounding and uncertainty quantification.

%\section{Reproducibility, Ethics, and AI use Statements}
%\textbf{Reproducibility Statement} The paper provides the methodological and experimental details required to reproduce the main results. We fully specify the ObsNODE architecture, continuous-time formulation, treatment modeling, training procedure, loss functions, optimization setup, and relevant hyperparameters. The datasets, including the synthetic cancer, semi-synthetic MIMIC-IV, and real-world sepsis data, as well as the construction of the semi-synthetic data, are described in detail. We further specify the evaluation metrics, prediction horizons, and experimental protocols, including repeated runs and aggregate reporting. These details enable independent reproduction of the experiments and verification of the main empirical findings.

%\textbf{Ethics Statement}
%This research involves no collection of new personal data. Experiments use synthetic, semi-synthetic, and publicly available clinical data (MIMIC-IV and AmsterdamUMCdb) in accordance with applicable access controls and data use agreements. No identifiable personal data are disclosed. We transparently report relevant assumptions, limitations, and potential sources of bias, and do not pursue harmful or deceptive applications of the proposed methods.

%\textbf{AI use Statement}
%Large language models (LLMs) were not used in the methodology, including model design, training, or evaluation. Their use was limited to writing and editing assistance and did not contribute to the scientific methodology, results, or conclusions.

\begin{ack}
	
\end{ack}

%\section*{References}
\bibliographystyle{rusnat}
\bibliography{refs_ObsNODE_new}

%\bibliographystyle{}

%%%%%%%%%%%%%%%%%%%%%%%%%%%%%%%%%%%%%%%%%%%%%%%%%%%%%%%%%%%%

\appendix

\section{Technical Appendix and Supplementary Material}

\subsection{Derivation of equation \ref{eq:outcome_ident}}\label{secapp:ident_derivation}
In the main text we derived only a discrete version of the adjustment formula \eqref{eq:outcome_ident} and here we provide the continuous time version.
Assume we have observed a treatment and outcome history $\vec{o}_{[0,t)}=\{\vec{a}_{[0,t)},\vec{y}_{[0,t)}\}$ before time~$t$ and also the outcome $\vec{Y}_{t}=\vec{y}_{t}$ at time $t$. We want to compute 
the probability density $P_{\vec{Y}_{t+s}(\vec{a}_{[t,t+s)})|\vec{Y}_t,\vec{O}_{[0,t)}}
(\vec{y}_{t+s}|\vec{y}_t,\vec{o}_{[0,t)})$ for the potential outcome $\vec{Y}_{t+s}\left(\vec{a}_{[t,t+s)}\right)$ under the hypothetical intervention path $\vec{a}_{[t,t+s)}$ applied in the time interval $[t,t+s)$.

Let us partition the data assimilation time interval $[0,t)$ into $K$ subintervals $[t_k,t_{k+1})$ with $t_0=0, t_K=t$ and $[0,t)=\cup_{k=1}^{K} [t_k,t_{k+1})$. Similarly, partition the prediction interval $[t,t+s)$ into $L$ subintervals with $[t_l,t_{l+1})$ with $t_{K+1}=t,t_{K+L}=t+s$ and $[t,t+s)=\cup_{l=K+1}^{K+L} [t_l,t_{l+1})$.

For each  $t_k \in [0,t)$ and $t_l \in [t,t+s)$ we identify the variables in the CFD~\eqref{eq:cfd_adjust} with variables in our model as
\begin{align*}
	A\leftarrowtail \vec{A}_{t},& \quad Y\leftarrowtail \vec{Y}_{t_l}, \quad  M \leftarrowtail \vec{Z}_{t_l}, \quad W \leftarrowtail \{\vec{Z}_t,\vec{Y}_t, \vec{A}_{t_k},\vec{Y}_{t_k}\}=\{\vec{Z}_t,\vec{Y}_t, \vec{O}_{t_k}\},
\end{align*} 
with $\vec{O}_{t_k}=\{\vec{A}_{t_k},\vec{Y}_{t_k}\}$. We can check graphically (replace $t_k$ by $0$ and $t_l$ by $t+s$ in Fig.~\ref{fig:fig1dags}) that $M=\vec{Z}_{t_l}$ fulfills the CFD~\ref{thm:CFD} with respect to $(A,Y)=(\vec{A}_{t},\vec{Y}_{t_l})$ with the conditioning variables $W=\{\vec{Z}_t,\vec{Y}_t,\vec{O}_{t_k}\}$.

In order to apply the CFD to sets of variables we define 
\begin{align*}
	\vec{A}_{0:K} := \{\vec{A}_{t_k}|t_k \in [0,t)\},\qquad \vec{Y}_{0:K} := \{\vec{Y}_{t_k}|t_k \in [0,t)\},\qquad 
	\vec{A}_{K+1:L} := \{\vec{A}_{t_l}|t_l \in [t,t+s)\}.
\end{align*}

Selecting the following sets  in Theorem~\ref{thm:CFD}
\begin{align*}
	A\leftarrowtail \vec{A}_{K+1:L},& \quad Y\leftarrowtail \vec{Y}_{t+s}, \quad  M \leftarrowtail \vec{Z}_{t+s}, \quad W \leftarrowtail \{\vec{Z}_t,\vec{Y}_t, \vec{A}_{0:K},\vec{Y}_{0:K}\}=\{\vec{Z}_t,\vec{Y}_t, \vec{O}_{0:K}\},
\end{align*} 
we see that $M=\vec{Z}_{t+s}$ fulfills the CFD relative to $(A,Y)=(\vec{A}_{K+1:L},\vec{Y}_{t+s})$ with the conditioning set $W=\{\vec{Z}_t,\vec{Y}_t, \vec{O}_{0:K}\}$. Applying the adjustment formula of the CFD~\eqref{eq:cfd_adjust} we have for the probability density of the potential outcome $\vec{Y}_{t+s}$ under the treatment sequence $\vec{a}_{K+1:L}$
\begin{align}\label{eq:full_adjustment_formula}
	P_{\vec{Y}_{t+s}(\vec{A}_{K+1:L})}\left(\vec{y}_{t+s}\right) &= \int_{\vec{z}_{t+s}}\int_{\vec{a}'_{K+1:L}}\int_{\vec{z}_t,\vec{y}_t, \vec{o}_{0:K}} 
	\left\{
	p(\vec{y}_{t+s}|\vec{a}'_{K+1:L},\vec{z}_{t+s},\vec{z}_{t},\vec{y}_t,\vec{o}_{0:K})\right.\\
	&\quad \times \left. p(\vec{a}'_{K+1:L}|\vec{z}_t,\vec{y}_t,\vec{o}_{0:K}) \, p(\vec{z}_{t+s}|\vec{a}_{K+1:L},\vec{z}_t,\vec{y}_t,\vec{o}_{0:K})
	\,p(\vec{z}_t,\vec{y}_t,\vec{o}_{0:K})\right\}.
\end{align}
Now we use the independence relations from assumptions \ref{ass:observation} and \ref{ass:latent}  
\begin{align*}
	P(\vec{y}_{t+s}|\vec{a}'_{K+1:L},\vec{z}_{t+s},\vec{z}_{t},\vec{y}_t,\vec{o}_{0:K}) &= p(\vec{y}_{t+s}|\vec{z}_{t+s})\\
	p(\vec{z}_{t+s}|\vec{a}_{K+1:L},\vec{z}_t,\vec{y}_t,\vec{o}_{0:K})	 &= p(\vec{z}_{t+s}|\vec{a}_{K+1:L},\vec{z}_t)
\end{align*}
and find for the probability density of the potential outcome
\begin{align*}
	P_{\vec{Y}_{t+s}(\vec{A}_{K+1:L})}&\left(\vec{y}_{t+s}\right) = \int_{\vec{z}_{t+s}}\int_{\vec{a}'_{K+1:L}}\int_{\vec{z}_t,\vec{y}_t, \vec{o}_{0:K}} 
	\left\{
	p(\vec{y}_{t+s}|\vec{z}_{t+s})\right.\\
	&\quad \times \left. p(\vec{a}'_{K+1:L}|\vec{z}_t,\vec{y}_t,\vec{o}_{0:K}) \, p(\vec{z}_{t+s}|\vec{a}_{K+1:L},\vec{z}_t)
	\,p(\vec{z}_t,\vec{y}_t,\vec{o}_{0:K})\right\}\\
	&=\int_{\vec{z}_{t+s}}\int_{\vec{z}_t,\vec{y}_t, \vec{o}_{0:K}}	
	p(\vec{y}_{t+s}|\vec{z}_{t+s})p(\vec{z}_{t+s}|\vec{a}_{K+1:L},\vec{z}_t)
	\,p(\vec{z}_t|\vec{y}_t,\vec{o}_{0:K})\,p(\vec{y}_t,\vec{o}_{0:K}).
\end{align*}
From this, we read
\begin{align}\label{eq:discrete_time_adjust}
	&P_{\vec{Y}_{t+s}(\vec{a}_{K+1:L})|\vec{Y}_t,\vec{O}_{0:K}}\left(\vec{y}_{t+s}|\vec{y}_t,\vec{o}_{0:K}\right) \\
	&\quad =\int_{\vec{z}_{t+s}}\int_{\vec{z}_t}	
	p(\vec{y}_{t+s}|\vec{z}_{t+s})p(\vec{z}_{t+s}|\vec{a}_{K+1:L},\vec{z}_t)
	\,p(\vec{z}_t|\vec{y}_t,\vec{o}_{0:K}).
\end{align}
This formula is valid for any partitions of the data assimilation interval $[0,t)$ and the forecast interval $[t,t+s)$. Making the partitions finer and finer by sending $K\to \infty$ and $L\to \infty$ allows us to replace $\vec{o}_{0:K}$ and by $\vec{o}_{[0,t)}$ and $\vec{a}_{K+1:L}$ by $\vec{a}_{[t,t+s)}$ and we obtain \eqref{eq:outcome_ident}. The model forecast is deterministic in \eqref{eq:node_ssm} and we have $p(\vec{z}_{t+s}|\vec{a}_{[t,t+s)},\vec{z}_t)= \delta\left(\vec{z}_{t+s}-\vec{z}_t-\int_t^{t+s}\,\vec{f}(\vec{z}_\tau,\vec{a}_\tau)\,d\tau\right)$.

Applying the continuum limit in \eqref{eq:discrete_time_adjust} only 
to the  hypothetical treatment trajectory ($L\to \infty$) we have
\begin{align*}
	&P_{\vec{Y}_{t+s}(\vec{a}_{[t,t+s)})|\vec{Y}_t,\vec{O}_{0:K}}\left(\vec{y}_{t+s}|\vec{y}_t,\vec{o}_{0:K}\right) \\
	&\quad =\int_{\vec{z}_{t+s}}\int_{\vec{z}_t}	
	p(\vec{y}_{t+s}|\vec{z}_{t+s})p(\vec{z}_{t+s}|\vec{a}_{[t,t+s)},\vec{z}_t)
	\,p(\vec{z}_t|\vec{y}_t,\vec{o}_{0:K}).
\end{align*}

If the discretization time points $t_k\in [0,t)$  for the data assimilation interval $[0,t)$ coincide with the time points for the observed data, we have an adjustment formula for discrete time observations and continuous time treatment trajectories. Here, $p(\vec{z}_t|\vec{y}_t,\vec{o}_{0:K})$ is the filtering distribution for the continuous time ODE model~\ref{eq:node_ssm} with discrete time observations~\citep{jazwinski_stochastic_2007}. 
\subsection{An adjustment formula for serially correlated hidden confounders}

\begin{figure}
\centering
\includegraphics[width=0.9\linewidth]{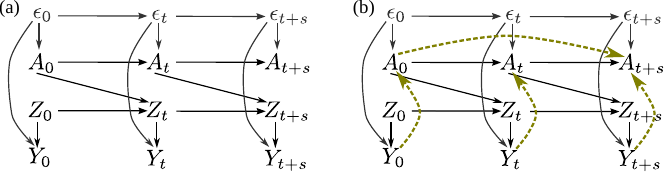}
\caption{(a) Temporal DAG for treatment $A$, observation $Y$, and latent state $Z$ at three consecutive time points. In contrast to Fig.~\ref{fig:fig1dags}, the unobserved process $\epsilon$ jointly affecting treatment assignment and observations exhibits temporal dependence. (b) Under a dynamic treatment regime, treatment additionally depends on the observed treatment--outcome history, as indicated by the green dashed arrows.}
\label{fig:Fig_crosscorrconf_DAGs}
\end{figure}

Assumption~\ref{ass:observation} is compatible with temporally independent unobserved disturbances $\vec{\epsilon}_t$ that affect both treatment assignment and contemporaneous observations, as shown in Fig.~\ref{fig:fig1dags}. If $\vec{\epsilon}_t$ is serially correlated, as illustrated in Fig.~\ref{fig:Fig_crosscorrconf_DAGs}, the observation at time $t+s$ may remain statistically dependent on the treatment trajectory even after conditioning on the latent state $\vec{Z}_{t+s}$. Consequently, the conditional independence
\begin{align*}
p(\vec{y}_{t+s}|
\vec{a}'_{K+1:L},\vec{z}_{t+s},\vec{z}_t,
\vec{y}_t,\vec{o}_{0:K})
=
p(\vec{y}_{t+s}|\vec{z}_{t+s})
\end{align*}
used in the preceding derivation no longer holds.

Nevertheless, provided that the CFD criterion remains satisfied, the potential-outcome distribution is still identified by the unsimplified adjustment formula~\ref{eq:full_adjustment_formula}. For a finite partition, conditioning the CFD adjustment formula on the observed history gives
\begin{align}
& P_{\vec{Y}_{t+s}(\vec{a}_{K+1:L})
|\vec{Y}_t,\vec{O}_{0:K}}
\left(
\vec{y}_{t+s}|\vec{y}_t,\vec{o}_{0:K}
\right)
\nonumber\\
&\quad =
\int_{\vec z_t}\int_{\vec z_{t+s}}
\int_{\vec a'_{K+1:L}}
\{
p\left(
\vec y_{t+s}|
\vec a'_{K+1:L},\vec z_{t+s},\vec z_t,
\vec y_t,\vec o_{0:K}
\right)
\nonumber\\
&\qquad\qquad\times
p\left(\vec a'_{K+1:L}| \vec z_t,\vec y_t,\vec o_{0:K}\right)
p\left(\vec z_{t+s}|\vec a_{K+1:L},\vec z_t\right)
p\left(\vec z_t|\vec y_t,\vec o_{0:K} \right)\}.
\label{eq:general_discrete_adjust}
\end{align}
Here, $\vec a_{K+1:L}$ denotes the hypothetical treatment sequence of interest, whereas $\vec a'_{K+1:L}$ is integrated with respect to its observational distribution.

In the continuum limit, $\vec a'_{K+1:L}$ becomes an alternative treatment trajectory $\vec a'_{[t,t+s)}$. Because the space of continuous-time trajectories is generally infinite-dimensional, the corresponding integration is expressed using the conditional path measure
\citep{kallenberg2021}
\begin{align*}
P\left(d
\vec a'_{[t,t+s)}
|\vec z_t,\vec y_t,\vec o_{[0,t)}
\right).
\end{align*}
The continuous-time adjustment formula is therefore
\begin{align}
& P_{\vec{Y}_{t+s}(\vec a_{[t,t+s)})
|\vec Y_t,\vec O_{[0,t)}}
\left(
\vec y_{t+s}|\vec y_t,\vec o_{[0,t)}
\right)
\nonumber\\
&\quad =
\int_{\vec z_t}\int_{\vec z_{t+s}}
\int_{\mathcal A_{[t,t+s)}}
\left\{
p\left(
\vec y_{t+s}|
\vec a'_{[t,t+s)},\vec z_{t+s},\vec z_t,
\vec y_t,\vec o_{[0,t)}
\right)\right.
\nonumber\\
&\qquad\qquad\times
\left.
P\left(
d\vec a'_{[t,t+s)}
|\vec z_t,\vec y_t,\vec o_{[0,t)}
\right)
p\left(
\vec z_{t+s}|
\vec a_{[t,t+s)},\vec z_t
\right)
p\left(
\vec z_t|\vec y_t,\vec o_{[0,t)}
\right)\right\}
\label{eq:general_continuous_adjust}
\end{align}
where $\mathcal A_{[t,t+s)}$ denotes the space of admissible treatment trajectories. Thus, observability still permits identification through the latent mediator, but serial dependence of the hidden confounder prevents the observational treatment-path distribution from being integrated out.

Equation~\ref{eq:general_continuous_adjust} could in principle be approximated by learning the conditional observational treatment process, sampling alternative treatment trajectories from this process, and averaging the conditional outcome density over these trajectories. Such an implementation additionally requires estimation of the complete treatment-path distribution and suitable path-space positivity.  However, its nonparametric estimation additionally requires path-space and mediator positivity, which may fail for deterministic latent dynamics. Practical implementations may therefore require stochastic latent dynamics or an augmented observable state that captures the temporally persistent confounding process.

The \mbox{ObsNODE} implementation considered in this work instead uses Assumption~\ref{ass:observation}, under which the outcome density is independent of $\vec a'_{[t,t+s)}$ given $\vec Z_{t+s}$. The path integral then equals one, and \eqref{eq:general_continuous_adjust} reduces to the tractable adjustment formula~\eqref{eq:outcome_ident}.

\subsection{Details on training the ObsNODE model}

The datasets were split into training, validation, and test sets and  feature-wise Z-score normalization was applied using statistics computed on the training set only. The normalized dataset is subsequently used for model training.

Each multivariate time series was segmented at a decision time point $t_c$ into two disjoint subsequences. The prefix segment consists of all observations prior to $t_c$ and is used to condition the recognition model. The model infers an initial state $\vec{z}_0$ for each individual trajectory using an an LSTM-based recognition network~\citep{rubanova_latent_2019}.

To handle missing observations, the recognition LSTM  is augmented with an explicit imputation mechanism.  The indicator function
\begin{align*}
\mathcal{I}_{tij} = \begin{cases}
	1: & y_{tij}\; \text{is observed}\\
	0: & y_{tij}\; \text{is missing}.
\end{cases}
\end{align*}
indicates whether the data for the $j-$th component of the $i-$th unit at time $t$ is observed or not,
The imputation layer is defined as the transformation
\begin{align*}
    {y}_{tij}
    =
    y_{tij} \cdot \mathcal{I}_{tij} 
    +
    b_{j} \cdot \left(1 - \mathcal{I}_{tij}\right),
\end{align*}
where $b_{j}$ are learnable parameters \citep{de_jong_deep_2019, wendland_generation_2022}. This formulation replaces missing entries with trainable values that are optimized jointly with the LSTM parameters in an end-to-end fashion.

The latent initial states $\vec{z}_0$ inferred by the LSTM, together with the control function $\vec{a}_t$ are subsequently passed as inputs to the Neural Differential Equation (NODE) model. The observations after the decision time point $t_c$ are used for loss computation and subsequent optimization. Model parameters are optimized using the Adam optimizer.

The ObsNODE model supports both direct long-horizon forecasting and a recursive  scheme, similar in spirit to ODE-RNN~\citep{rubanova_latent_2019}. In the long-horizon setting, the state estimated by the LSTM recognition model is used to generate a forecast over the entire prediction window. In the recursive setting, forecasts are produced iteratively: at each step, the ODE model generates a short-term prediction, which is appended to the observation and treatment history sequence observed up to $t_c$. Conditioned on this augmented sequence, the LSTM updates the latent state, which then serves as the initial condition for the next ODE rollout. This process repeats until the full horizon is covered, without using any ground-truth observations beyond $t_c$.
We find empirically that augmenting the sequence with intermediate predictions can improve stability for longer forecasting horizons in some settings.

\subsubsection{Code availability}
ObsNODE is implemented in PyTorch. Code is available at \url{https://github.com/JenniferJaschob/ObsNODE.git}. 
The repository includes scripts for data preprocessing, simulation, model training, and evaluation, together with configuration files specifying all hyperparameters used in the experiments.

%

%\subsection{Additional Results}\label{subsec:additional_results}

%Tables~\ref{tab:add_res_tumorvolumen} and  \ref{tab:add_res_semisynmimiciv} provide supplementary results from the experimental stability analysis. 

%\subsubsection{Details of Results Semi-Synthetic Data}

%\subsubsection{Details of Results real-world data}
%Figure~\ref{fig:resmimic4} and ~\ref{fig:resamsterdamUMCdb} shows the heatmaps of the mean RMSE across different assimilation horizons for the MIMIC-IV dataset and external validation on AmsterdamUMCdb dataset.

\subsection{Details for Computational Experiments}
\subsubsection{Details for Synthetic Data Simulation}\label{subsec: tumor_growth_simulator}
The synthetic data set is simulated from a dynamic model for tumor volume $V(t)$ (in $cm^3$) \citep{geng_prediction_2017,bica_estimating_2020} and net body weight  $W^{net}(t)$ (in $kg$)  in response to a dynamic dosing trajectory for chemotherapy  $C(t)$ and radiotherapy  $d(t)$ combination treatment~\citep{bellazzi_dynamic_2025}:
\begin{subequations}
	\begin{align}
		\frac{dV\left(t\right)}{dt} &= \left( \rho \log \left( \frac{K}{V \left( t \right)} \right) - \beta_{c} C \left( t \right) - \left( \alpha_{r} d \left( t \right) + \beta_{r} d \left( t \right)^{2} \right) + \epsilon(t) \right) V \left( t \right)\\
		\frac{d W^{net} \left( t \right) }{dt} &= k_{g} W^{net} \left( t \right) \left(1-\frac{W^{net} \left( t \right)}{G} \right) - \beta_{wc} C \left( t \right) - \alpha_{wr} d \left( t \right) - \lambda V \left( t \right).
	\end{align}
\end{subequations}
The tumor volume is assumed to follow a Gompertz growth function, which is inhibited by the chemotherapy and radiotherapy dosing. The Gaussian zero mean white noise process $\epsilon(t)$ renders the tumour volume a stochastic variable.

Net body weight $W^{net}(t)$ is modeled as a logistic growth process, modified by tumor volume and therapy-induced weight loss~\citep{bellazzi_dynamic_2025}. Net body weight $W^{net}(t) = W(t) - \rho_{d} V(t)$ is defined as body weight $W(t)$ minus the tumor weight, with 
$\rho_{d}$ denoting the tumor density. 
The parameter $G$ corresponds to the saturation level, which in this case is set to the initial weight. The parameters $\beta_{wc}$, $\alpha_{wr}$, and $\lambda$ determine the weight loss contributions of chemotherapy, radiotherapy, and disease impact, respectively. 

Treatment doses are modeled as time dependent random variables 
\begin{subequations} \label{eqapp:dose_prob}
\begin{align}
	C(t)&=  C_{max} \, \sigma \left(\frac{\gamma_c \,\alpha_c^{dose}}{D_{max}}(\overline{D}(t)-\delta_c)\right)\\
	d(t)&=d_{max} \, \sigma \left(\frac{\gamma_r\,\alpha_r^{dose}}{D_{max}}(\overline{D}(t)-\delta_r)\right),
\end{align}
\end{subequations}
where $\sigma$ is the softmax function and $C_{max} = 14\, \nicefrac{\text{mg}}{\text{m}^{3}}$ and $d_{max} =3\, \text{Gy}$ are the maximum doses for for chemo- and radiotherapy, respectively. The prescribed treatment doses depend on the  outcome via the mean tumor diameter $\overline{D}(t)$ over the last 15 days \citep{melnychuk_causal_2022}. The parameter $D_{max}=13\,cm$ is the maximum tumor diameter and we set $\delta_c=\delta_r=\frac{D_{max}}{2}$. The  degree of time dependent confounding is given by the  parameters $\gamma_c$ and $\gamma_r$, with range $\gamma =\gamma_c=\gamma_r \in [1,8]$.
The values of $\alpha_c^{dose}$ and $\alpha_c^{dose}$ are sampled uniformly in the range from $1 \text{cm}$ to $4 \text{cm}$ to simulate variability in individual dosing decisions. The IGC-Net model can only process binary treatments indicators $\vec{a}_t$ and we interprete $C(t)/C_{max}$ and $d(t)/d_{max}$ for IGC-Net as probabilities for the Bernoulli-variables of treatment for chemo- and radiotherapy, respectively.

We simulate 3,000 patients and use 1,000 patients each for training, validation, and testing. Chemotherapy dose $C(t)$ and radiotherapy dose $d(t)$ was adjusted in monthly treatment cycles by sampling from~\eqref{eqapp:dose_prob}. 
To model patient variability, we sampled patient specific parameters from Gaussian  distributions, see Table~\ref{tab:parameter_cancer_simulation}.
\begin{table}[h!]
	\vskip 0.15in 
	\begin{center}
		\begin{small}
			\begin{sc}
				%\begin{tabular}{|l|l|l|l|l|}
                \begin{tabular}{lllll}
					\hline
					\textbf{Model} & \textbf{Variable} & \textbf{Parameter} & \textbf{Distribution} & \textbf{Distribution Value ($\mu$, $\sigma$)} \\
					\hline
					Tumor & Growth parameter & $\rho$ & Normal & $7 \times 10^{-5},\ 0.00723$ \\
					Tumor & Carrying capacity & $K$ & Constant & $30$ \\
					Tumor & Radio cell kill ($\alpha$) & $\alpha_r$ & Normal & $0.0398,\ 0.168$ \\
					Tumor & Radio cell kill ($\beta$) & $\beta_r$ & - & set $\alpha/\beta = 10$ \\
					Tumor & Chemo cell kill & $\beta_c$ & Normal & $0.028,\ 0.0007$ \\
					Tumor & Noise & $\epsilon_{vt}$ & Normal & $0,\ 0.01$ \\
					\hline
					Weight&  Growth parameter & $\rho_w$ & Normal & $14 \times 10^{-5},\ 10^{-5}$ \\
					Weight & Carrying capacity & $K_w$ & Constant & Initial weight \\
					Weight & Radio influence & $\alpha_{wr}$ & Normal & $0.004125,\ 10^{-4}$ \\
					Weight & Chemo influence & $\beta_{wc}$ & Normal & $0.001775,\ 2 \times 10^{-4}$ \\
					Weight & Cancer influence & $\lambda$ & Normal & $31 \times 10^{-5},\ 15 \times 10^{-6}$ \\
					Weight&  Noise & $\epsilon_{wt}$ & Normal & $0,\ 0.0015$ \\
					\hline
				\end{tabular}
			\end{sc}
			\caption{Parameter of the cancer growth simulation model \citep{bellazzi_dynamic_2025}.}
			\label{tab:parameter_cancer_simulation}
		\end{small}
	\end{center}
	\vskip -0.1in
\end{table}

We generated five counterfactual treatment scenarios ($CF_1$–$CF_5$). For $CF_1$ and $CF_2$ , we sampled two different treatment sequences by independently and uniformly selecting one of four predefined treatment application levels at each time step (chemotherapy: $\{$0, 3, 5, 7$\}$, radiotherapy: $\{$0, 1, 2, 3$\}$). These sampled dose sequences were then fixed and applied deterministically to all patients after the intervention time. These values represent the applied values, while the effective dose is modeled via the pharmacokinetic drug decay process. $CF_1$ and $CF_2$ therefore represent randomly assigned alternative treatment sequences. $CF_3$ denotes a strategy consisting of maximum radiotherapy without chemotherapy, while $CF_4$ denotes a strategy consisting of maximum chemotherapy without radiotherapy. $CF_5$ corresponds to no treatment.

\subsubsection{Details for Semi-Synthetic Data Simulation}
\label{subsec:semi_synthetic_data_simulation}
To generate semi-synthetic data  \citep{schulam_reliable_2018} we have modified the approach from \citep{melnychuk_causal_2022}. We use the sepsis patient cohort (3557 patients) from the MIMIC-IV dataset \citep{johnson_mimic-iv_2023} (see also  \ref{subsec:mimicIV_data}) and select three vital signs as time dependent hidden confounders $\vec{\epsilon}_t$. Then, we simulate untreated outcomes $j\in \{1,\ldots,d_y\}$ as
\begin{align}\label{eqapp:semi_untreatedOutcomes}
	\tilde{Y}_t^{j,(i)} =  \alpha_s^j\, \textit{B-spline(t)} +\alpha_g^j\, g^{j,(i)}(t) + \alpha_\phi^j\, \phi_Y^j(\vec{\epsilon}_t^{(i)}) + \eta_t^{j,(i)}
\end{align}
with noise $\eta_t^{j,(i)} \sim \mathcal{N}(0,\, 0.005^2)$ and weight coefficients $\alpha_s^j,  \alpha_g^j$ and $ \alpha_\phi^j$.
The term $\textit{B-spline}(t)$ represents a mixture of three cubic B-splines.  The function $g^{j,(i)}$ is independently sampled for each patient $i$ from a Gaussian process with a Matérn kernel, and $\phi_Y^j$ is sampled via random Fourier features (RFF) \citep{hensman_variational_2018}.  

Synthetic binary treatments $\vec{A}_t=(A_t^1,\ldots,A_t^{d_a})\in \{0,1\}^{d_a}$ with confounding are simulated via  
\begin{align*}
	A_t^l &\sim \text{Bernoulli}(P_{A_t^l}),\qquad P_{A_t^l} = \sigma \left( \gamma_A^l \, \bar{Y}_{t-1}^{A,l}  + \gamma_{\epsilon}^l \Phi_{\epsilon}^l(\vec{\epsilon}_t) + b_l \right), 
\end{align*}
where $\sigma$ is the sigmoid function. The treatment probability $P_{A_t^l}$ for the $l-$th treatment component $A_t^l$ depends on the outcome history, represented by an average over a subset of the previous treated outcomes $\bar{Y}_{t-1}^{A,l}$~\citep{melnychuk_causal_2022}. Hidden confounding of the treatment is represented via the function $\Phi_{\epsilon}^l$ sampled from a RFF approximation of a Gaussian process. The parameters $\gamma_A^l$ and $\gamma_{\epsilon}^l$ control the confounding strength and $b_l$ is a bias 	 parameter.  

The time dependent treatment effect 
\begin{align}
	E^{j}_t &= \sum_{k=t-w^l}^t 
	\frac{\min_{l=1,\dots,d_a} \mathbf{1}_{[A_k^l=1]} P_{A_k^l} \beta_{lj}}
	{(w^l - k)^2}.
\end{align}
is bounded by the maximum effect size $\beta_{lj}\in \{0,\beta\}$ of treatment component $l$ on outcome $j$. It is either zero for outcomes $j$ not affected by treatment $l$ or equals the constant $\beta$. Finally, the treated outcomes are obtained from the untreated outcomes~\eqref{eqapp:semi_untreatedOutcomes} by adding the treatment effects

\begin{align}
Y_t^{j,(i)}=\tilde{Y}_t^{j,(i)} + E^{j}_t\,.
\end{align}

Parameters for the semi-synthetic sepsis data are given in Table~\ref{tab:parameters_semi_synth_sepsis}.

\begin{table}[htbp]
	\centering

	\begin{tabular}{lll}
    %\begin{tabular}{|l|l|l|}
		\hline
		\textbf{Variable} & \textbf{Parameter} & \textbf{Value} \\
		\hline
		Outcome & $\alpha$ & 2 \\
		Outcome & $\alpha_g$ & 0.5 \\
		Outcome & $\alpha_e$ & 1 \\
		Outcome & $\nu$ & 20 \\
		Outcome & Exogeneous Variables & Creatinine, Bilirubin, ALT \\
		\hline
		Treatment 1; 2 & $\gamma_z$ & 0.3; 0.3 \\
		Treatment 1; 2 & $\gamma_0$ & 0.3; 0.1 \\
		Treatment 1; 2 & bias & -2; -2 \\
		Treatment 1; 2 & $\nu$ & 20; 20 \\
		Treatment 1; 2 & Confounding Variables & Creatinine; Bilirubin, ALT \\
		\hline
	\end{tabular}
    \caption{\textbf{Parameters of the data-simulation model for semi-synthetic data~\citep{bellazzi_dynamic_2025}.}  }
	\label{tab:parameters_semi_synth_sepsis}
\end{table}

We generated five counterfactual treatment scenarios $CF_k$ for $k\in{1,\dots,5}$ by modifying the treatment assignment mechanism during simulation. $CF_1$ corresponds to a randomized treatment policy, with treatment probabilities sampled randomly. For $CF_2$--$CF_5$, treatment probabilities are deterministically set to one or zero, corresponding to maximal or minimal treatment, respectively. $CF_2$ implements a strategy with maximum first treatment and no subsequent treatment, whereas $CF_3$ represents a strategy with no first treatment and maximum second treatment. $CF_4$ corresponds to a no-treatment strategy, while $CF_5$ represents a strategy with maximum treatment for all time points after the decision time point $t_c$.

\newpage
\subsubsection{Details on Experiments with Real-World Data}
\label{subsec:mimicIV_data}
\textbf{MIMICIV sepsis cohort:} We use the MIMIC-IV electronic health record dataset collected at Beth Israel Deaconess Medical Center, which contains data from approximately 450,000 hospital admissions \citep{johnson_mimic-iv_2023}. The dataset includes demographic information, laboratory values, vital signs, medications, and diagnoses.

For our analysis we focus on sepsis patients identified using the OpenSep pipeline according to the Sepsis-3 criteria \citep{hofford_opensep_2022}. Sepsis cases were defined by an increase in the Sequential Organ Failure Assessment (SOFA) score of at least two points within 48 hours before or 24 hours after a suspected infection.

The SOFA score is used as the clinical outcome variable. It summarizes dysfunction in six organ systems (respiratory, cardiovascular, central nervous system, coagulation, liver, and kidney), each scored from 0 to 4 according to the severity of dysfunction \citep{vincent_sofa_1996}. An overview of the subscores is given in Table~\ref{tab:sofa}.

\begin{table}[tbh]	
	%\vskip 0.15in
	\begin{center}
		\begin{scriptsize}
			\begin{sc}
				\begin{tabular}{lllllll}
                %\begin{tabular}{|l|l|l|l|l|l|l|}
					\hline
					\textbf{Organ} & \textbf{Parameter} &\textbf{Unit}& \textbf{1} & \textbf{2} & \textbf{3} & \textbf{4} \\
					\hline
					Respiration &$PaO_2/FiO_2$ & $mmHg$& $<$400 & $<$300 & $<$200 & $<$100 \\
					
					\hline
					Coagulation & Platelets & $\mu l$ &$<$150.000 & $<$100.00 & $<$50.000 & $<$20.000\\
					\hline
					\makecell{Liver} & Bilirubin & mg/dl &1.2-1.9 & 2.0-5.9 & 6.0-11.9 & $>$12.0\\
					\hline
					\makecell{Cardio-\\vascular} & \makecell{MAP  or\\
						Vasopressors}& \makecell{mmHG or\\ $\mu g/kg / min$}&MAP $<$ 70 & \makecell{Dopam. $\leq$ 5 \\or dobutam.} & \makecell{Dopam. $>$ 5 \\or epineph. $\leq$0.1 \\or norepin. $\leq$ 0.1} & \makecell{Dopam. $>$ 15 \\or epineph. $>$ 0.1 \\or norepin. $>$ 0.1} \\
					\hline
					\makecell{Central \\Nervous\\ System} & GCS& &13-14 & 10-12 & 6-9 & $<$6 \\
					\hline
					Renal & \makecell{Creatinine } &\makecell{mg/dl \\or ml/d} &1.2-1.9 & 2.0-3.4 & \makecell{3.5-4.9 \\or urin $<$ 500}  & $>$5.9\\
					\hline
				\end{tabular}
			\end{sc}
			
			\caption{\textbf{Overview of the SOFA score subscores} \citep{wendland_optimal_2024} Each column corresponds to a subscore (0–4) reflecting the severity of dysfunction in the central nervous, cardiovascular, respiratory, coagulation, hepatic, and renal systems. Subscores of 0 indicate normal organ function as defined by the respective criteria.}
			\label{tab:sofa}
		\end{scriptsize}
	\end{center}
	\vskip -0.1in
\end{table}

We consider patients treated with the antibiotics vancomycin, ceftriaxone, and piperacillin-tazobactam. To assess potential treatment side effects we additionally predict laboratory indicators of organ toxicity, including creatinine (kidney injury) as well as total bilirubin and alanine aminotransferase (ALT) (liver injury).

Data preprocessing follows the OptAB pipeline \citep{wendland_optimal_2024}. The cohort is restricted to ICU patients, and antibiotic treatments are encoded as a vector of three binary variables indicating administration of vancomycin, ceftriaxone, and piperacillin-tazobactam. 

Time-dependent covariates were selected from all available dynamic variables by excluding variables recorded in fewer than 50\% of patients and retaining those with an absolute Spearman correlation of at least 0.3 with the SOFA score during the first 12 hours. The resulting covariates are listed in Table~\ref{tab:time_dependent_covariates_mimiciv} .

To capture information about measurement patterns, cumulative missingness indicators were added as additional time-dependent covariates. Static covariates include sex, age, height, and weight at admission.

The final dataset contains 3,556 patients. We use 80\% of the data for training and 20\% for testing, with the training set further split into 80\% training and 20\% validation data.
\begin{table}[bht]	
	\vskip 0.15in
	\begin{center}
		\begin{small}
			\begin{sc}
				\begin{tabular}{ll}
                %\begin{tabular}{|l|l|}
					\hline
					\textbf{Variable} & \textbf{Unit} \\
					\hline
					SOFA score & - \\
					Alanine aminotransferase & IU/L \\
					Anion gap & mEq/L \\
					Bicarbonate & mEq/L \\
					Total bilirubin & mg/dl \\
					Blood urea nitrogen (BUN) & mg/dl \\
					Creatinine & mg/dl \\
					Diastolic blood pressure & mmHg \\
					Platelet count & k/$\mu$L \\
					Red cell distribution width (RDW) & \% \\
					Systolic blood pressure & mmHg \\
					\hline
				\end{tabular}
			\end{sc}
			\caption{Time-dependent covariates from MIMIC-IV-sepsis dataset}
			\label{tab:time_dependent_covariates_mimiciv}
		\end{small}
	\end{center}
	\vskip -0.1in
\end{table}

\textbf{AmsterdamUMCdb sepsis cohort:} For external validation, we evaluated the performance of ObsNODE, trained on MIMIC-IV, in the independent AmsterdamUMCdb cohort~\citep{thoral_sharing_2021}. AmsterdamUMCdb comprises Electronic Health Records (EHR) data from 23,106 ICU admissions at Amsterdam University Medical Center between 2003 and 2016. The distribution of administered antibiotics differed substantially between AmsterdamUMCdb and MIMIC-IV, potentially reflecting differences in patient populations, pathogen prevalence, or local treatment practices. Among 6,285 patients with sepsis in AmsterdamUMCdb, 3,374 received Ceftriaxone, 3,058 Cefotaxime, and 1,718 erythromycin. For external validation, we restricted the cohort to patients treated with antibiotics  Vancomycin, Ceftriaxone, or Piperacillin/Tazobactam. This resulted in an external test set of 399 patients, comprising 364 treated with Ceftriaxone and 35 with vancomycin; no patients treated with Piperacillin/Tazobactam.

The two cohorts differed substantially in both SOFA score distributions and the availability of laboratory measurements. The mean SOFA score among patients with sepsis was higher in AmsterdamUMCdb than in MIMIC-IV (6.8 to 4.43). Laboratory measurements contributing to the SOFA score were also more frequently available in AmsterdamUMCdb. In the AmsterdamUMCdb test set, 23.3\% of patients had no recorded total bilirubin measurement and 16.8\% had no recorded alanine aminotransferase (ALT) measurement, compared with 56.8\% and 56.1\%, respectively, in MIMIC-IV. These differences indicate substantial heterogeneity in laboratory testing and documentation practices, which may contribute to the observed distribution shift and affect model robustness and generalizability across cohorts.

The AmsterdamUMCdb data were preprocessed following the OptAB pipeline \citep{wendland_optimal_2024}, using the same preprocessing procedures as applied to the MIMIC-IV dataset to ensure consistency between the training and external validation cohorts.

\textbf{Data availability:} The public MIMIC-IV dataset \citep{johnson_mimic-iv_2023, goldberger_physiobank_2000} is available under \url{https://physionet.org/content/mimiciv/2.2/} and is shared by the data owners after reasonable request. 

AmsterdamUMCdb dataset \citep{thoral_sharing_2021}  is available under \url{https://amsterdammedicaldatascience.nl/amsterdamumcdb/}. The data is shared by the data owners after reasonable request including the supervision of a practising intensivist.

\newpage
\subsection{Hyperparameter Tuning}
\label{sec:hyperopt}
\subsubsection{Hyperparameter Tuning of ObsNODE}
For hyperparameter optimization of the ObsNODE model, we use Optuna with the Tree-structured Parzen Estimator (TPE). Table \ref{tab:hyperparameter_ObsNODE} summarizes the search space (a) and the selected configurations for three datasets: the synthetic cancer dataset, the semi-synthetic MIMIC-IV dataset, and the MIMIC-IV sepsis dataset (b). While the parameters of the NODE component can, in principle, be tuned separately for each neural network in the observation normal form \eqref{eq:triang_nf}, we keep them identical across all components in our experiments.

\begin{table}[htb]
	
	\vskip 0.15in
	\begin{center}
		\begin{small}
			\begin{sc}
				\begin{tabular}{lcc}
                %\begin{tabular}{|l|c|c|}
					\hline
					Search Space/Data set & \makecell{synthetic \\cancer} & \makecell{(semi-synthetic)\\ MIMIC-IV} \\
					\hline
					\hline
					Latent space dimension & $[d_y,..,3\,d_y]$ & $[d_y,..,3\,d_y]$\\
					\hline
					Batch size    & \{32,64,...,256\} & [100,250,500,750,1000]\\
					\hline
					Learning rate & [0.001, 0.0001, 0.00001] & [0.001, 0.0001, 0.00001]  \\
					\hline
					Hidden dim Observer    & [32,64,128,256] & [32,64,128,256]   \\
					\hline
					Hidden dimension NODE    & [32,64,128,256] & [32,64,128,256]   \\
					\hline
					Number of layers NODE &  \{1,2,...,10\} & \{1,2,...,10\} \\
					\hline
					Activation function NODE &  [leakyrelu, tanh, sigmoid] & [leakyrelu, tanh, sigmoid]\\
					\hline
				\end{tabular}
				\subcaption{Search space for ObsNODE hyperparameters.}
				\begin{tabular}{lccc}
					\hline
					Hyperparameter/Data set & \makecell{synthetic\\ cancer} & \makecell{semi-synthetic\\ MIMIC-IV} & MIMIC-IV   \\
					\hline
					Latent space dimension = $d_y \times m$ & $2\times 2$ & $1\times 2$  &  $11\times 3$\\
					Batch size    & 64 & 100 &  500\\
					Learning rate & 0.001 & 0.0001 &      0.0001 \\
					Hidden dimension Observer    & 128 & 128 &   256   \\
					Hidden dimension NODE    & 128 & 128  &  256 \\
					Number of layers NODE &  5 & 6 &  7\\
					Activation function NODE &   leakyrelu & leakyrelu & leakyrelu\\
					\hline
				\end{tabular}
				\subcaption{Optimized hyperparameter values for ObsNODE.}
			\end{sc}
			\caption{\textbf{Hyperparameter Tuning of the ObsNODE Model.} Table (a) illustrates the hyperparameter search space, while Table (b) reports the optimal hyperparameter configurations for experiments conducted on the synthetic cancer dataset, the semi-synthetic MIMIC-IV sepsis dataset, and the MIMIC-IV sepsis dataset.}
			\label{tab:hyperparameter_ObsNODE}
		\end{small}
	\end{center}
	\vskip -0.1in
\end{table}

\newpage
\subsubsection{Hyperparameter Tuning of OptAB and DoseAI}
For hyperparameter tuning of the OptAB and DoseAI models, we use Optuna with the Tree-structured Parzen Estimator (TPE). Table \ref{tab:hyperparameter_OptAB_DoseAI} summarizes the search space (a) and the resulting configurations (b) for both models. DoseAI can handle continuous treatment doses and is applied to the synthetic cancer dataset, while OptAB is restricted to categorial treatments and used for the semi-synthetic dataset based on MIMIC-IV and the real world sepsis dataset from MIMIC-IV. The decoder optimization is initialized with the final hyperparameter values obtained for the encoder.
\begin{table}[htb]
	
	\vskip 0.15in
	\begin{center}
		\begin{small}
			\begin{sc}
				\begin{tabular}{lcc}
                %\begin{tabular}{|l|c|c|}
					\hline
					Search Space/Data set & \makecell{synthetic \\cancer} & \makecell{(semi-synthetic)\\ MIMIC-IV} \\
					\hline
					\hline
					Dimension of the latent state & \{1, 2, ..., 30\} & \{1, 2, ..., 30\}   \\
					\hline
					batch size &\makecell{\{16, 32, 64, 125,\\250, 500, 1000\}\\ }& \makecell{\{100, 200, 500,\\1000, 2000\}\\}  \\
					\hline
					Max. number of NCDE units  & \{1, 2, ..., 1000\} & \{1, 2, ..., 1000\}   \\
					\hline
					Learning rate  & [0.0001, 0.01] & [0.0001, 0.01]    \\
					\hline
					Activation function of F  & \makecell{\{leakyrelu, tanh,\\sigmoid, identity\}\\} & \makecell{\{leakyrelu, tanh,\\sigmoid, identity\} \\}   \\
					\hline
					Number of NCDE layers  & \{1, 2, ..., 20\}& \{1, 2, ..., 20\}   \\
					\hline
					Activation function of $h_{\alpha}$ & \makecell{\{leakyrelu, tanh,\\sigmoid, identity\}\\} & \makecell{\{leakyrelu, tanh,\\sigmoid, identity\} \\}  \\
					\hline
					Number of layers of h & \{1, 2, ..., 1000\} &  \{1, 2, ..., 1000\}  \\
					\hline
					Max. number of units of h & \{1, 2, ..., 6\} & \{1, 2, ..., 6\}    \\
					\hline	
				\end{tabular}
				\subcaption{Search space for OptAB and DoseAI  hyperparameters.}
				\begin{tabular}{lcccr}
					\hline
					Hyperparameter/Data set & \makecell{synthetic\\ cancer} & \makecell{semi-synthetic\\ MIMIC-IV} & MIMIC-IV   \\
					\hline
					encoder & & &  \\
					\hline
					Dimension of the latent state & 16& 7 & 17 \\  
					batch size & 250& 1000 & 500 \\ 
					Max. number of NCDE units  & 578& 174 & 33 \\ 
					Learning rate  & 0.0042& 0.0016&  0.0051 \\
					Activation function of F  & leakyrelu&leakyrelu & tanh  \\ 
					Number of NCDE layers  & 2& 14 & 15 \\ 
					Activation function of $h_{\alpha}$ & tanh& leakyrelu & tanh \\
					Number of layers of h & 128& 275 & 128 \\ 
					Max. number of units of h & 4& 4 & 1 \\  
					\hline
					decoder & & &  \\
					\hline
					Dimension of the latent state  & 22 & 25 & 17 \\
					batch size  & 125 & 1000 & 1000 \\
					Max. number of NCDE units  & 802 &   825 & 33 \\
					Learning rate  & 0.0016 & 0.0007 & 0.0051 \\
					Activation function of F   & leakyrelu & identity & tanh \\
					Number of NCDE layers  & 13 & 9 & 15 \\
					Activation function of $h_{\alpha}$ & leakyrelu & leakyrelu & tanh \\ 
					Number of layers of h & 798 & 403 &  128\\
					Max. number of units of h & 1 & 1 & 1 \\ 
					\hline
				\end{tabular}
				\subcaption{Optimized hyperparameter values for OptAB and DoseAI Models.}
			\end{sc}
			\caption{\textbf{Hyperparameter Tuning of the OptAB and DoseAI Models.} Table (a) presents the hyperparameter search space, while Table (b) reports the optimal hyperparameter configurations. DoseAI is used for training on the synthetic cancer dataset, whereas OptAB is applied to the semi-synthetic data set and the real world sepsis dataset from MIMIC-IV.}
			\label{tab:hyperparameter_OptAB_DoseAI}
		\end{small}
	\end{center}
	\vskip -0.1in
\end{table}

\newpage
\subsubsection{Hyperparameter Tuning of IGC-Net}
For hyperparameter tuning of the IGC-Net model, we employ the Hydra implementation of the random grid search strategy for computational effciency reasons. Table \ref{tab:hyperparameter_ObsNODE} presents the hyperparameter search space (a) and the selected hyperparameter configuration (b) of the IGC-Net model.
\begin{table}[htb]
	
	\vskip 0.15in
	\begin{center}
		\begin{small}
			\begin{sc}
				\begin{tabular}{lccc}
                %\begin{tabular}{|l|c|c|c|}
					\hline
					Search Space/ Data set & \makecell{synthetic \\cancer} & \makecell{semi-synthetic\\ MIMIC-IV} &  MIMIC-IV  \\
					\hline
					\hline
					Transformer blocks  & [1,2] &1 & 1 \\
					\hline
					Learning rate ($\eta$) & \makecell{[0.01, 0.001,\\ 0.0001]} & \makecell{[0.001, 0.0001, \\ 0.0001]}  &\makecell{[0.001, \\0.0001]} \\
					\hline
					Minibatch size &[64, 128, 256] & \makecell{[32,64,128, \\ 256,512]} & [32,64]\\
					\hline
					Attention heads  & 1 &[1,2,3] & [2,3]\\
					\hline
					Transformer units &  \makecell{[$1d_g, 2d_{g},$\\
						$3d_{g}, 4d_{g}$]}  &  \makecell{[$1d_{g}, 2d_{g},$ \\$3d_{g}, 4d_{g}$]} & \makecell{[$0.5d_{g}, 1d_{g}$]}  \\
					\hline
					Hidden representation size &  \makecell{[$0.5d_{g}, 1d_{g}, $ \\$ 2d_{g}, 3d_{g}, 4d_{g}$]} &  \makecell{[$1d_{g},2d_{g},$ \\$  3d_{g}, 4d_{g}$]}&  \makecell{[$0.5d_{g}, 1d_{g},$ \\$2d_{g}$]}\\
					\hline
					Feed-forward hidden units  &  \makecell{[$0.5d_z , 1d_z ,$\\ $ 2d_z, 3d_z , 4d_z$]} & \makecell{$ [1d_z , 2d_z ,$\\ $3d_z , 4d_z$] }&  \makecell{[$0.5d_z , 1d_z ,$\\ $2d_z$]}\\
					\hline
					Sequential dropout rate  & [0.1, 0.2] & [0.1, 0.2] & [0.1, 0.2]\\
					\hline
					Max positional encoding  & 15 & 30 & 15\\
					\hline
					Number of epochs & 50 & 100 & 100\\
					\hline
				\end{tabular}
				\subcaption{Search space for IGC-Net hyperparameters.}
				\begin{tabular}{lcccr}
					\hline
					Hyperparameter/Data set & \makecell{synthetic\\ cancer} & \makecell{semi-synthetic\\ MIMIC-IV} & MIMIC-IV   \\
					\hline
					Transformer blocks  & 2 & 2& 1\\
					Learning rate  & 0.01 & 0.01 &   0.001    \\
					Minibatch size & 64 & 64 &64\\
					Attention heads   & 0.1 & 1 & 2\\
					Transformer units  & 32 & 24& 30\\
					Hidden representation size & 64 & 72& 30\\
					Feed-forward hidden units  & 32 & 18 & 30\\
					Sequential dropout rate & 0.1 & 0.1 &0.1 \\
					Max positional encoding  & 15 & 30 & 15\\
					Number of epochs  & 50 & 100 & 100\\
					\hline
				\end{tabular}
				\subcaption{Optimized hyperparameter values for IGC-Net.}
				
			\end{sc}
		\caption{\textbf{Hyperparameter Tuning of the IGC-Net Model.} Table (a) presents the hyperparameter search space, while Table (b) reports the optimal hyperparameter configurations of the IGC-Net model for experiments conducted on the synthetic cancer dataset, the semi-synthetic MIMIC-IV dataset, and the real world MIMIC-IV sepsis dataset. The overall input dimension is given by $d_{g} = d_y + d_x + d_a$, and $d_z$ denotes the hidden representation size of the IGC-Net model.}
			\label{tab:hyperparameter_IGC-Net}
		\end{small}
	\end{center}
	\vskip -0.1in
\end{table}

\newpage
\subsubsection{Hyperparameter Tuning of  SCIP-Net}
For hyperparameter tuning of the SCIP-Net model for the synthetic cancer data and semi-synthetic  data, we employed the Hydra–Optuna implementation of the Tree-structured Parzen Estimator (TPE). For the real sepsis data, we tuned the SCIP-Net model using a random grid search. Hyperparameter optimization with Optuna using the Tree-structured Parzen Estimator (TPE) did not yield improved or optimal parameter configurations.

Table \ref{tab:hyperparameter_SCIP-Net} summarizes the hyperparameter search space (a) and the selected configurations (b) for the synthetic cancer dataset, the semi-synthetic MIMIC-IV sepsis dataset, and the MIMIC-IV sepsis dataset.
\begin{table}[htb]
	
	\vskip 0.15in
	\begin{center}
		\begin{small}
			\begin{sc}
				\begin{tabular}{lcc}
                %\begin{tabular}{|l|c|c|}
					\hline
					
					Search Space\ Model & \makecell{Weight network, \\Treatment network,\\ Encoder} & Decoder \\
					\hline
					\hline
					
					Neural CDE hidden layers   &  [1,2] & [1,2]\\
					\hline
					Learning rate & \makecell{[0.01,0.001,\\ 0.0001, 0.0001]}& \makecell{[0.01,0.001,\\ 0.0001, 0.0001]} \\
					\hline
					Minibatch size     &[64,128,256]  & [256,512,1024]  \\
					\hline
					Neural CDE hidden units & [0.5,1,2,3,4]$d_{g}$ &  [1,2,3,4,8,16]$d_{g}$\\
					\hline
					Neural CDE dropout rate    &  [0.1,0.2] & [0.1,0.2]\\
					\hline
					Max. gradient norm    &  [0.5,1,2] & [0.5,1.0,2.0,4.0]  \\
					\hline
					Epochs & 50 & 50\\
					
					\hline
					%\multicolumn{3}{c}{\textbf{(A) Search Space}}\\
				\end{tabular}
				\subcaption{Search space for SCIP-Net hyperparameters.}
				\begin{tabular}{lcccc}
					\\
					\toprule

					Parameter / Model  & \makecell{Weight \\ network} & \makecell{Treatment \\network}  &Encoder & Decoder \\
					\hline
					Synthetic cancer & & & & \\
					\hline
					Neural CDE hidden layers   & 2 & 1 & 1 &2 \\
					Learning rate & 0.0001 & 0.01 & 0.01 & 0.001\\
					Minibatch size     & 128 & 64 & 64 & 256 \\
					Neural CDE hidden units & 8 & 12 & 8 & 32\\
					Neural CDE dropout rate    & 0.1 & 0.1& 0.1& 0.1\\
					Max. gradient norm    &  0.5 & 0.5 & 0.5 & 1 \\
					%Epochs & 50 & 50 & 50 & 50\\
					
					\hline
					Semi-synthetic MIMIC-IV  & & & & \\
					\hline
					Neural CDE hidden layers   & 1 & 2 & 2 &2 \\
					Learning rate & 0.01 & 0.01 & 0.01 & 0.0001\\
					Minibatch size     & 64 & 64 & 256 & 256 \\
					Neural CDE hidden units & 9 & 12 & 6 & 12\\
					Neural CDE dropout rate    & 0.1 & 0.1& 0.2& 0.2\\
					Max. gradient norm    &  0.5 & 0.5 & 2 & 1 \\
					%Number of epochs & 50 & 50 & 50 & 50\\
					
					\hline
					
					MIMIC-IV  & & & & \\
					\hline
					Neural CDE hidden layers   & 2 & 2 & 1 & 2\\
					Learning rate & 0.01 & 0.01 & 0.01 & 0.01\\
					Minibatch size     & 256 & 256 & 256 & 512 \\
					Neural CDE hidden units & 42 & 42& 42 & 84\\
					Neural CDE dropout rate    & 0.2  & 0.1 & 0.1 & 0.1\\
					Max. gradient norm    & 0.5  & 2.0 & 2.0 & 0.5 \\
					%Number of epochs & 50 & 50 & 50 & 50 \\
					\hline
				\end{tabular}
			%\end{sc}
			\subcaption{Optimized hyperparameter values for SCIP-Net.}
            \end{sc}
			\caption{\textbf{Hyperparameter Tuning of the SCIP-Net Model.} Table (a) presents the hyperparameter search space. Table (b) reports the optimal configurations for the synthetic cancer dataset and the semi-synthetic MIMIC-IV sepsis dataset and the real world MIMIC-IV sepsis dataset.}
			\label{tab:hyperparameter_SCIP-Net}
		\end{small}
	\end{center}
	\vskip -0.1in
\end{table}

\newpage
\subsubsection{Hyperparameter Tuning of NODE-Model}
As a baseline model, we implemented a Neural Ordinary Differential Equation (NODE) model using the same encoder-LSTM architecture as employed for the ObsNODE model. In contrast to the ObsNODE formulation, the latent dynamics are modeled using a standard latent Neural ODE without incorporating an observability normal form. The latent dimension was chosen to be identical to that of the ObsNODE model to ensure a fair comparison between both approaches.

For the hyperparameter optimization of the NODE model, we employed Optuna with the Tree-structured Parzen Estimator (TPE) sampler. The considered search space and the resulting optimal hyperparameter configuration are summarized in Table~\ref{tab:hyperparameter_NODE}.
\begin{table}[htb]
	\vskip 0.15in
	\begin{center}
		\begin{small}
			\begin{sc}
				\begin{tabular}{lcc}
					\hline
					Search Space/ Data set & \makecell{synthetic \\cancer} & \makecell{semi-synthetic\\ MIMIC-IV}  \\
					\hline
					
					\hline
					Batch size    & \{32,64,...,256\} & [100,250,500,750,1000] \\
					\hline
					Learning rate & [0.001, 0.0001, 0.00001] & [0.001, 0.0001, 0.00001] \\
					\hline
					Hidden dimension Observer    & [32,64,128,256] & [32,64,128,256] \\
					\hline
					Hidden dimension NODE    & [32,64,128,256] &  [32,64,128,256] \\
					\hline
					Number of layers NODE &  \{1,2,...,10\} & \{1,2,...,10\}\\
					\hline
					Activation function NODE &  [leakyrelu, tanh, sigmoid] & [leakyrelu, tanh, sigmoid]\\
					\hline
				\end{tabular}
                \subcaption{Search space for NODE hyperparameters.}

                \begin{tabular}{lcc}
					\hline
					Hyperparameter/ Data set & \makecell{synthetic \\cancer} & \makecell{semi-synthetic\\ MIMIC-IV} \\
					\hline
					
					\hline
					Batch size    &  96 & 100\\
					\hline
					Learning rate &  0.001 & 0.001 \\
					\hline
					Hidden dimension Observer    &  256 & 64 \\
					\hline
					Hidden dimension NODE    &   64 & 64\\
					\hline
					Number of layers NODE &   8 & 3\\
					\hline
					Activation function NODE &  sigmoid & leakyrelu\\
					\hline
				\end{tabular}
                \subcaption{Optimized hyperparameter values for NODE.}
                \end{sc}
			\caption{\textbf{Hyperparameter tuning of the NODE model.} Hyperparameter search space (a) and optimal configuration (b) for the NODE model on the synthetic cancer and semi-synthetic MIMICIV-sepsis dataset.}
			\label{tab:hyperparameter_NODE}
		\end{small}
	\end{center}
	\vskip -0.1in
\end{table}

\subsection{Computational efficiency}
For training and hyperparameter optimization of ObsNODE we utilized two Nvidia A100 GPUs, each with 40 GB of memory. The same setting was used for the baseline models DoseAI, OptAB and SCIP-Net. For IGC-Net, we used a NVIDIA H100 GPU with 96 GB of memory.

Training of ObsNODE for the optimal hyperparameter values on a single Nvidia A100 GPU took 140 min for the synthetic cancer data, 49 min for the semi-synthetic  data and 65 min for the real world sepsis data.

This is similar to the training times required for the baseline models IGC-Net and SCIP-Net. Training OptAB and DoseAI required between 40 and 74 hours, depending on the data set.

\subsection{ Declaration of LLM usage}
The core method development in this research does not involve LLMs as any important original, or non-standard components.

\end{document}